\documentclass[10pt,conference,compsocconf,final]{IEEEtran}

\IEEEoverridecommandlockouts

\def\BibTeX{{\rm B\kern-.05em{\sc i\kern-.025em b}\kern-.08em
    T\kern-.1667em\lower.7ex\hbox{E}\kern-.125emX}}

\usepackage{enumitem}

\newcommand{\dnnscope}{\textit{Ben\allowbreak an\allowbreak za}\xspace}

\usepackage[export]{adjustbox}
\usepackage{pstricks}
\usepackage{amsmath,amssymb,amsfonts}
\usepackage{graphicx}
\usepackage{xcolor}

\usepackage{amsmath}

\usepackage[breakable, theorems, skins]{tcolorbox}
\tcbset{enhanced}

\usepackage[bookmarks=false,bookmarksnumbered=true,
pdfpagemode={UseOutlines},plainpages=false,pdfpagelabels=true,
colorlinks=true,linkcolor={black},citecolor={black},urlcolor={black},breaklinks=true]{hyperref}

\usepackage{microtype}
\usepackage{verbatim}
\usepackage{booktabs} %
\usepackage{soul}
\usepackage{amsthm}
\usepackage{amsmath}
\usepackage{fancyhdr}
\usepackage{xspace}
\usepackage{multirow}
\usepackage{array}
\usepackage{makecell}
\usepackage{rotating}
\usepackage{filecontents}
\usepackage{pgfplots}
\usepackage{pgfplotstable}
\usepackage{tikz}
\usepackage[normalem]{ulem}
\usepackage[eulergreek]{sansmath}

\usepackage{xargs}                      %
\usepackage{amssymb}%
\usepackage{pifont}%
\usepackage{environ}
\usepackage{tabularx}%
\usepackage{arydshln}

\newcommand{\yes}{{\color{black} \ding{51}}}
\newcommand{\no}{{\color{black} \ding{55}}}

\newcommand{\cmmnt}[1]{\ignorespaces}

\definecolor{myred}{rgb}{0.843137,0.188235,0.152941}
\definecolor{myblack}{rgb}{0.27451,0.32549,0.384314}
\definecolor{mygreen}{rgb}{0.301961,0.686275,0.290196}
\definecolor{myyellow}{rgb}{0.996078,0.878431,0.564706}
\definecolor{myblue}{rgb}{0.568627,0.74902,0.858824}

\pgfplotsset{compat=newest,}
\usepgfplotslibrary{external}

\definecolor{lightyellow}{RGB}{255, 250, 236}
\definecolor{textdark}{RGB}{100, 52, 20}
\definecolor{borderorange}{RGB}{253, 129, 36}
\definecolor{lightgray}{RGB}{214, 214, 214}
\definecolor{countrygray}{RGB}{153, 153, 153}
\definecolor{circlered}{RGB}{176, 0, 29}

\pgfplotsset{every axis/.style={scale only axis}}

\pgfplotsset{%
    ,tick label style = {font=\small\sansmath\sffamily\footnotesize} %
    ,every axis label = {font=\small\sansmath\sffamily\footnotesize}
    ,legend style = {font=\tiny\sansmath\sffamily\tiny}
    ,label style = {font=\sansmath\sffamily\footnotesize}
}

\usepgfplotslibrary{colorbrewer}
\pgfplotsset{cycle list/Dark2-8}
\pgfplotsset{cycle list/RdYlBu-6}
\pgfplotsset{cycle list/Set1}
\pgfplotsset{cycle list/Paired}

\newcommand{\ignore}[1]{}

\newtcolorbox{observationbox}[1][]
{
    breakable,
    left=1pt,
    right=1pt,
    top=1pt,
    bottom=1pt,
    colback=gray!20,
    colframe=black,
    width=\dimexpr\columnwidth\relax,
    enlarge left by=0mm,
    boxsep=5pt,
    arc=0pt,outer arc=0pt,
    #1
}

\makeatletter
\newsavebox{\measure@tikzpicture}
\NewEnviron{scaletikzpicturetowidth}[1]{%
  \def\tikz@width{#1}%
  \def\tikzscale{1}\begin{lrbox}{\measure@tikzpicture}%
  \BODY
  \end{lrbox}%
  \pgfmathparse{#1/\wd\measure@tikzpicture}%
  \edef\tikzscale{\pgfmathresult}%
  \BODY
}

\newtcbox\questionbox{hbox, on line, colback=black, enhanced, frame hidden, boxrule=0pt, 
    top=-2pt, bottom=-2pt, right=-2pt, left=-2pt, rounded corners, arc=2pt}
    
\DeclareRobustCommand\question[1]{\tikz[baseline=(char.base)]{
            \node[rounded corners,white,draw,fill=black,line width=1pt,rounded corners=2pt,inner sep=1.5pt,anchor=base] (char) {$\mathcal{Q}$\normalfont\sffamily\textsf{#1}};}\nolinebreak\ignorespacesafterend\hspace{-2pt}}

\DeclareRobustCommand*\circledwhitered[1]{\tikz[baseline=(char.base)]{
            \node[shape=circle,line width=0.5mm,circlered,draw,text=black,inner sep=0.5pt,anchor=base] (char) {\normalfont\sffamily\bfseries\footnotesize{#1}};}\nolinebreak\ignorespacesafterend\hspace{-2pt}}
\DeclareRobustCommand*\circledwhite[1]{\tikz[baseline=(char.base)]{
            \node[shape=circle,line width=0.5mm,draw,inner sep=0.5pt,anchor=base] (char) {\normalfont\sffamily\bfseries\footnotesize{#1}};}\nolinebreak\ignorespacesafterend\hspace{-2pt}}

\usepackage{setspace}

\title{\dnnscope{}: Automatic $\mu$Benchmark Generation to Compute ``Lower-bound'' Latency and Inform Optimizations of Deep Learning Models on GPUs}


\author{
\IEEEauthorblockN{Cheng Li\textsuperscript{*}, Abdul Dakkak\textsuperscript{*}}
\IEEEauthorblockA{University of Illinois Urbana-Champaign\\
Urbana, USA\\
\{cli99, dakkak\}@illinois.edu}
\and
\IEEEauthorblockN{Jinjun Xiong}
\IEEEauthorblockA{IBM T. J. Watson Research Center\\
Yorktown Heights, USA\\
jinjun@us.ibm.com}
\and 
\IEEEauthorblockN{Wen-mei Hwu}
\IEEEauthorblockA{University of Illinois Urbana-Champaign\\
Urbana, USA\\
w-hwu@illinois.edu}
}

\begin{document}

\maketitle

\footnotetext[1]{The two authors contributed equally to this paper.}

\begin{abstract}
As Deep Learning (DL) models have been increasingly used in latency-sensitive applications, there has been a growing interest in improving their response time. 
An important venue for such improvement is to profile the execution of these models and characterize their performance to identify possible optimization opportunities.
However, the current 
profiling tools 
lack the highly desired abilities to %
characterize ideal performance, identify sources of inefficiency, and quantify the benefits of 
potential optimizations.
Such deficiencies have led to slow characterization/optimization cycles 
that cannot keep up with the fast pace at which new DL models are introduced.

We propose \dnnscope, a sustainable and extensible benchmarking and analysis design that speeds up the characterization/optimization cycle of DL models on GPUs.
\dnnscope consists of four major components: a model processor that parses models into an internal representation, a configurable benchmark generator that automatically generates micro-benchmarks given a set of models, a database of benchmark results, and an analyzer that computes the ``lower-bound'' latency of DL models using the benchmark data and informs optimizations of model execution.
The ``lower-bound'' latency metric estimates the ideal model execution on a GPU system and serves as the 
basis for identifying optimization opportunities in frameworks or system libraries.
We used \dnnscope to evaluate $30$ ONNX models in MXNet, ONNX Runtime, and PyTorch on $7$ GPUs ranging from Kepler to the latest Turing, and identified optimizations in parallel layer execution, cuDNN convolution algorithm selection, framework inefficiency, layer fusion, and using Tensor Cores.

\end{abstract}

% \begin{IEEEkeywords}
% Deep Learning; GPU; Performance; Benchmarking; Optimization;
% \end{IEEEkeywords}

\section{Introduction}\label{sec:intro}

The past few years have seen a spur of deep learning (DL) innovations.
These innovations span from DL models to software stack optimizations (e.g. frameworks such as MXNet or PyTorch, libraries such as cuDNN or MKL-DNN) and hardware stack improvements (e.g. CPU, GPU, FPGA).
Among all the innovations, however, DL models are the most rapidly evolving and prolific.
This is true in both academia~\cite{dean2018new} and industry~\cite{hazelwood2018applied}, where models are tweaked and introduced on a weekly, daily, or even hourly basis.

Both industry and academia have invested heavily in developing benchmarks to characterize DL models and systems~\cite{mlperf,aimatrix,deepbench,dawnbench,deep500}.
Characterization is followed by optimizations to improve the model performance.
However, there is currently a gap between the benchmarking results and possible optimizations to perform.
Researchers use profilers, such as  nvprof~\cite{nvprof}, Nsight~\cite{nsight}, and VTune~\cite{vtune}, to profile and get low-level GPU and CPU information.
With ample knowledge of how models execute and utilize system resources, researchers manually identify bottlenecks and inefficiencies within model execution using the profilers.
Researchers then make hypotheses of solutions, and try out different ideas to optimize the model execution --- which may or may not pan out.
This manual and ad-hoc process requires a lot of effort 
and expertise
and slows down the turnaround time for model optimization and system tuning.

\begin{figure}[t]
	\centering
	\includegraphics[width=0.48\textwidth]{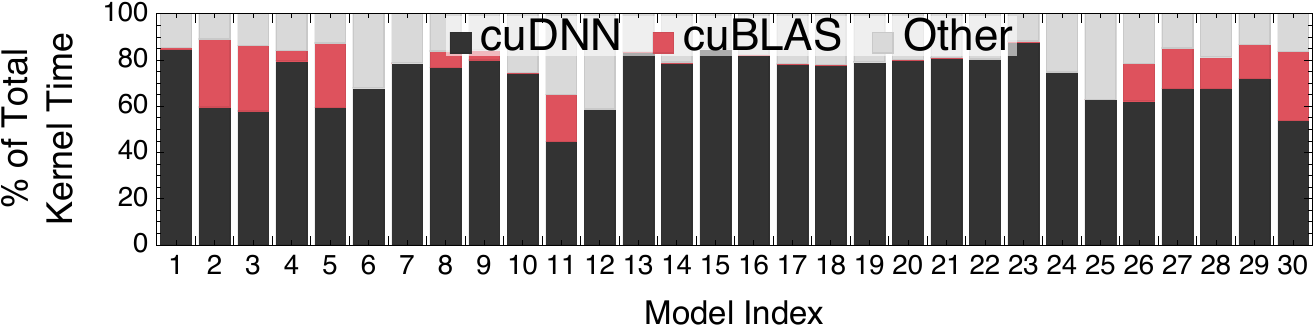}
	\caption{
		The GPU kernel time breakdown for all $30$ models (listed in Table~\ref{tab:models}) on Tesla\_V100 (Table~\ref{tab:systems}) using batch size 1.
		Both cuDNN and cuBLAS invoke child GPU kernel(s) asynchronously.
		Therefore, we measure the time of the kernels launched by the cuDNN and cuBLAS APIs rather than the time of the APIs themselves for accurate characterization of latencies.
	}
	\label{fig:perf_breakdown}
\end{figure}

Thus there is a need for a systematic DL benchmarking and subsequent analysis design that can guide researchers to potential optimization opportunities and assess hypothetical execution scenarios.
Since for GPUs model execution latency is determined by the hardware, framework, and system libraries (primarily cuDNN~\cite{cudnn} and cuBLAS~\cite{cublas} for DL),
answers to the following questions are highly desired by researchers:
\question{1} what is  the potential latency speedup if optimizations are performed?
\question{2} Are independent layers executed in parallel?
\question{3} Are convolution layers using the optimal convolution algorithms?
\question{4} Are there any inefficiencies or unexpected behavior in a framework?
Does the execution \question{5} fuse layers or \question{6} leverage Tensor Cores, and what are the benefits?
We motivate our design by answering these $6$ questions, while ensuring the sustainability and extensibility of the design. %

To answer these questions, we first propose a new benchmarking metric:
\textit{``lower-bound'' latency}.
The ``lower-bound'' latency estimates the ideal latency of a DL model given a software and hardware stack, and is based on the following observations:
(1) DL models are executed as layers in frameworks and thus layers form the performance building blocks of DL models. (2) Frameworks delegate execution of common layers to either cuDNN or cuBLAS (shown in Figure~\ref{fig:perf_breakdown}).
The ``lower-bound'' latency is defined in terms of the latencies of the cuDNN and cuBLAS API functions corresponding to the model layers (framework overhead and memory transfers are ignored).
We refine the ``lower-bound'' latency and define it under \textit{sequential execution mode} (all layers are executed sequentially) and \textit{parallel execution mode} (data-independent layers are executed asynchronously).

This paper presents \dnnscope{} (pronounced bonanza) --- a sustainable and extensible benchmarking and analysis design.
\dnnscope consists of a set of modular components: (1) a model processor to process input ONNX models into a set of \textit{unique layers} (layers are considered the same if they have the same layer type, shape, and parameters), (2) a benchmark generator to automatically generate parameterized cuDNN and cuBLAS micro-benchmarks from the unique layers, (3) a performance database to store historical benchmark results, and (4) an analyzer to compute the ``lower-bound'' latency of DL models and inform potential optimizations (\question{1-6}~).

\dnnscope{} is architected to be sustainable.
The benchmarking workflow of \dnnscope{} is highly automated and minimizes the benchmark development and maintenance effort.
\dnnscope{} uses the observation that DL models have repeated layers (i.e. non-unique) within and across models to decrease the time to benchmark.
When a new model is introduced, only the new, un-benchmarked layers
(not in the performance database) need to be benchmarked.
Although the focus of the paper is on NVIDIA GPUs using cuDNN and cuBLAS, the design proposed is extensible and users can incorporate other benchmark runtimes that target other software libraries or hardware such as: frameworks' API or MKL-DNN for CPUs.

In summary, this paper makes the following contributions:
\begin{itemize}[nosep,leftmargin=0.5em,labelwidth=*,align=left]
	\item We propose a ``lower-bound'' latency metric for DL models based on the observation that the latency of a DL model is bounded by the latencies of the cuDNN and cuBLAS API calls corresponding to the model layers.
	    The ``lower-bound'' latency metric estimates the ideal latency of a model given a specific GPU hardware and software stack.
	\item We present \dnnscope, a novel benchmarking and %analyzing design that that automatically generates
	 analysis system designed to automatically generate micro-benchmarks given a set of models;
	 compute their ``lower-bound'' latencies using the benchmark data; and
	inform optimizations
	of their 
	execution on GPUs.
    \dnnscope{} is sustainable and extensible to cope with the fast evolution of DL innovations.
	\item Using \dnnscope{}, we characterized the ``lower-bound'' latencies of $30$ ONNX models (shown in Table~\ref{tab:models}) using MXNet, ONNX Runtime, and PyTorch on $7$ systems (shown in Table~\ref{tab:systems}).
	We performed a comprehensive ``lower-bound'' latency analysis as we vary the model, execution mode, batch size, and system.
	E.g., when using parallel execution mode, up to $2.87\times$(with a geometric mean of $1.32\times$ across models) latency speedup could be made to MXNet using batch size $1$ on the \texttt{Tesla\_V100} system.
	\item We 
	%further 
	identified optimization opportunities through \dnnscope in cuDNN convolution algorithm selection (up to $1.32\times$ geometric mean speedup across models), inefficiencies within MXNet (up to $1.15\times$ speedup across models) and PyTorch (up to $2.3\times$ speedup using batch size $1$) frameworks, and layer fusion and Tensor Cores (up to $1.09\times$ and $1.72\times$ speedup for \texttt{ResNet50-v1} respectively).
	%We evaluated the above optimizations jointly and 
	We further demonstrated that when performed jointly, these optimizations 
	%get 
	achieve up to $1.95\times$ speedup for \texttt{ResNet50-v1} across systems and batch sizes.
% 	\item The usages and evaluation of \dnnscope{} are posted to \websiteurl{} for public inspection.
	      %
\end{itemize}

\begin{figure*}
	\centering
	\setlength{\abovecaptionskip}{-10pt}
	\setlength{\belowcaptionskip}{-10pt}
	\includegraphics[width=\textwidth]{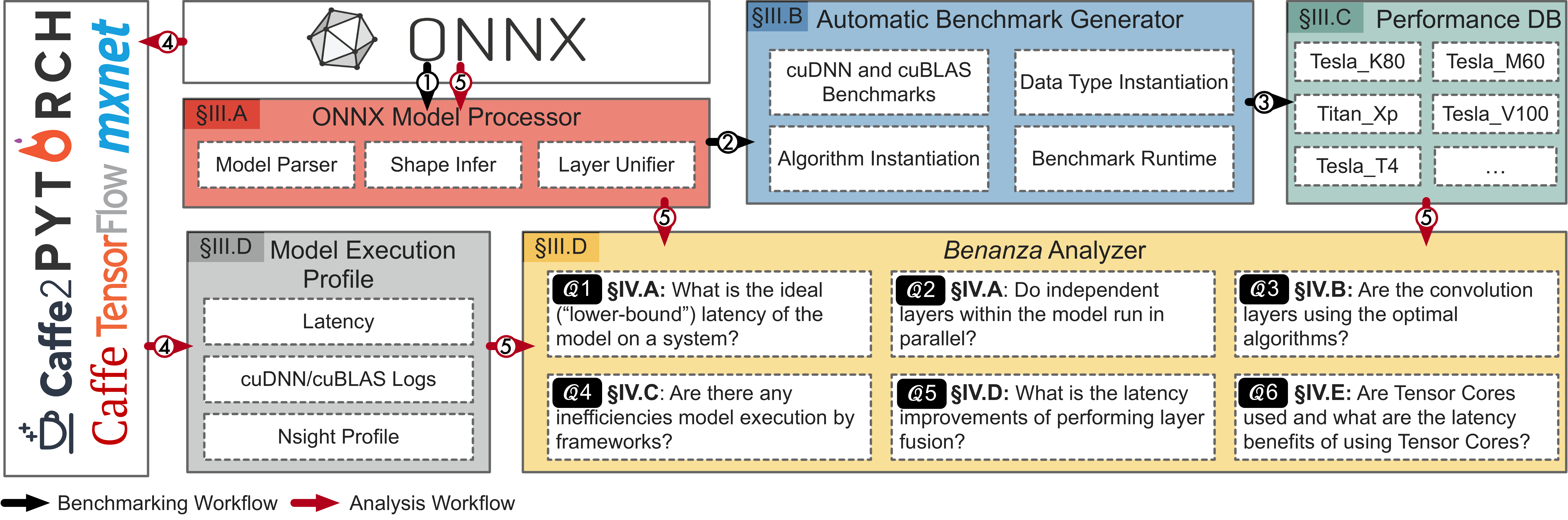}
	\caption{
		The \dnnscope design and workflow.
	}
	\label{fig:dnn_scope_flow}
\end{figure*}

\section{Background and Motivation}\label{sec:motivation}

\subsection{DL Model Execution and ONNX Format}\label{sec:onnx}

A DL model is an execution graph where each vertex is a layer operator (e.g. convolution, activation, normalization, pooling, or softmax).
These layer operators (or \textit{layers} for short) are functions defined by a DL framework.
A framework executes a model by traversing the model graph in topological order and enqueuing the layers into an execution queue.
Although sequential evaluation is always valid, frameworks strive to execute data-independent layers within the queue in parallel.
Through execution scheduling, a framework can overlap communication with computation, run two data-independent layers in parallel, etc.
Regardless of the execution strategy, however, layer execution latency is the limiting factor for model execution.
Therefore, layers are not only the building blocks by which developer define models, but are also the atomic components that define a model's performance characteristics.

Each framework provides its own API, layer definition semantics,  model storage format, and model executing strategy.
To increase interoperability between frameworks, there has been concerted effort~\cite{onnx,nnef} to standardize layer definitions  and model exchange format.
A leading effort is the Open Neural Network Exchange Format (ONNX), which has wide industry and framework backing.
Frameworks such as Caffe2, CNTK, MXNet, Paddle, PyTorch, and TensorRT readily support ONNX, and converters exist for other frameworks such as Caffe and TensorFlow.
To perform a fair comparison between frameworks (by evaluating them using the same ONNX model), and more importantly, to make \dnnscope framework-agnostic, we choose ONNX as the model input format for \dnnscope.
ONNX hosts all their models publicly~\cite{onnxzoo} and, we select $30$ vision models out of the $32$ models available at the time of writing for evaluation (the $2$ models not selected are non-vision models).
The selected models cover an array of tasks and are listed in Table~\ref{tab:models}.
We refer to these models by their IDs throughout the paper.

\begin{table}
	\centering
	\caption{ The $30$ ONNX models used are vision models which encompass image classification (\textbf{IC}), object detection (\textbf{OD}), face recognition (\textbf{FR}), emotion recognition (\textbf{ER}), semantic segmentation (\textbf{SS}), or hand digit recognition (\textbf{HR}) tasks.
	}
	\resizebox{0.41\textwidth}{!}{%
		\begin{tabular}{rlcrrr} \toprule
			\centering%
			\textbf{\thead{ID}} & \textbf{\thead{Name}}                                      & \textbf{\thead{Task}} & \textbf{\thead{MACs}} & \textbf{\thead{\# Layers}} & \textbf{\thead{Year}} \\ \midrule
			1                   & Arcface~\cite{DBLP:journals/corr/abs-1801-07698}           & FR                    & 12.08G                & 412                        & 2018                  \\
			2                   & BVLC-Alexnet~\cite{alexnet}                                & IC                    & 656M                  & 24                         & 2012                  \\
			3                   & BVLC-Caffenet~\cite{alexnet}                               & IC                    & 721M                  & 24                         & 2012                  \\
			4                   & BVLC-Googlenet~\cite{googlenet}                            & IC                    & 1.59G                 & 143                        & 2014                  \\
			5                   & BVLC-RCNN-ILSVRC13~\cite{DBLP:journals/corr/GirshickDDM13} & IC                    & 718M                  & 23                         & 2013                  \\
			6                   & Densenet-121~\cite{DBLP:journals/corr/HuangLW16a}          & IC                    & 2.87G                 & 910                        & 2016                  \\
			7                   & DUC~\cite{DBLP:journals/corr/WangCYLHHC17}                 & SS                    & 34.94G                & 355                        & 2017                  \\
			8                   & Emotion Ferplus~\cite{DBLP:journals/corr/BarsoumZCZ16}     & ER                    & 877M                  & 52                         & 2016                  \\
			9                   & Inception-v1~\cite{DBLP:journals/corr/IoffeS15}            & IC                    & 1.44G                 & 144                        & 2015                  \\
			10                  & Inception-v2~\cite{DBLP:journals/corr/SzegedyVISW15}       & IC                    & 2.03G                 & 509                        & 2015                  \\
			11                  & LeNet~\cite{lecun1998gradient}                             & HR                    & 796K                  & 12                         & 2010                  \\
			12                  & MobileNet-v2~\cite{DBLP:journals/corr/HowardZCKWWAA17}     & IC                    & 437M                  & 155                        & 2017                  \\
			13                  & Resnet18-v1~\cite{DBLP:journals/corr/HeZRS15}              & IC                    & 1.82G                 & 69                         & 2015                  \\
			14                  & Resnet18-v2~\cite{DBLP:journals/corr/HeZR016}              & IC                    & 1.82G                 & 69                         & 2016                  \\
			15                  & Resnet34-v1~\cite{DBLP:journals/corr/HeZRS15}              & IC                    & 3.67G                 & 125                        & 2015                  \\
			16                  & Resnet34-v2~\cite{DBLP:journals/corr/HeZR016}              & IC                    & 3.67G                 & 125                        & 2016                  \\
			17                  & Resnet50-v1~\cite{DBLP:journals/corr/HeZRS15}              & IC                    & 3.87G                 & 175                        & 2015                  \\
			18                  & Resnet50-v2~\cite{DBLP:journals/corr/HeZR016}              & IC                    & 4.10G                 & 174                        & 2016                  \\
			19                  & Resnet101-v1~\cite{DBLP:journals/corr/HeZRS15}             & IC                    & 7.58G                 & 345                        & 2015                  \\
			20                  & Resnet101-v2~\cite{DBLP:journals/corr/HeZR016}             & IC                    & 7.81G                 & 344                        & 2016                  \\
			21                  & Resnet152-v1~\cite{DBLP:journals/corr/HeZRS15}             & IC                    & 11.30G                & 515                        & 2015                  \\
			22                  & Resnet152-v2~\cite{DBLP:journals/corr/HeZR016}             & IC                    & 11.53G                & 514                        & 2016                  \\
			23                  & Shufflenet~\cite{DBLP:journals/corr/ZhangZLS17}            & IC                    & 127M                  & 203                        & 2015                  \\
			24                  & Squeezenet-v1.1~\cite{DBLP:journals/corr/IandolaMAHDK16}   & IC                    & 352M                  & 66                         & 2016                  \\
			25                  & Tiny Yolo-v2~\cite{DBLP:journals/corr/RedmonF16}           & OD                    & 3.13G                 & 32                         & 2016                  \\
			26                  & Vgg16-BN~\cite{DBLP:journals/corr/SimonyanZ14a}            & IC                    & 15.38G                & 54                         & 2014                  \\
			27                  & Vgg16~\cite{DBLP:journals/corr/SimonyanZ14a}               & IC                    & 15.38G                & 41                         & 2014                  \\
			28                  & Vgg19-bn~\cite{DBLP:journals/corr/SimonyanZ14a}            & IC                    & 19.55G                & 63                         & 2014                  \\
			29                  & Vgg19~\cite{DBLP:journals/corr/SimonyanZ14a}               & IC                    & 19.55G                & 47                         & 2014                  \\
			30                  & Zfnet512~\cite{DBLP:journals/corr/ZeilerF13}               & IC                    & 1.48G                 & 22                         & 2013                  \\
			\bottomrule
		\end{tabular}%
	}
	\label{tab:models}
\end{table}

\begin{table}
	\centering
	\caption{
		Eleven layer types are supported by cuDNN
		and two layer types are supported by cuBLAS. %
		Each API may have auxiliary functions to setup its arguments (e.g. \texttt{cudnnSetTensor4dDescriptor} to specify a tensor's dimensions and \texttt{cudnnSetConvolution2dDescriptor} to configure the convolution API).
		The convolution, RNN, and GEMM APIs have Tensor Core support.
	}
	\resizebox{0.45\textwidth}{!}{%
		\begin{tabular}{l l c} \toprule
			\centering%
			\textbf{\thead{Layer Type}} & \textbf{\thead{cuDNN / cuBLAS API}}              & \textbf{\thead{\shortstack{Tensor Core} \\ {Support}}}
			\\ \midrule
			Convolution                 & \texttt{cudnnConvolutionForward}                 & \yes                                    \\
			Activation                  & \texttt{cudnnActivationForward}                  & \no                                     \\
			BatchNorm                   & \texttt{cudnnBatchNormalizationForwardInference} & \no                                     \\
			Conv+Bias+Activation        & \texttt{cudnnConvolutionBiasActivationForward}   & \yes                                    \\
			RNN                         & \texttt{cudnnRNNForwardInference}                & \yes                                    \\
			Dropout                     & \texttt{cudnnDropoutForward}                     & \no                                     \\
			Pooling                     & \texttt{cudnnPoolingForward}                     & \no                                     \\
			Softmax                     & \texttt{cudnnSoftmaxForward}                     & \no                                     \\
			Add                         & \texttt{cudnnAddTensor}                          & \no                                     \\
			Element-wise                & \texttt{cudnnOpTensor}                           & \no                                     \\
			Rescale                     & \texttt{cudnnScaleTensor}                        & \no                                     \\ \hdashline
			GEMM                        & \texttt{cublas*Gemm / cublasGemmEx}              & \yes                                    \\
			GEMV                        & \texttt{cublasSgemv}                             & \no                                     \\
			\bottomrule
		\end{tabular}
	}
	\label{tab:cudnnapi}
\end{table}

\subsection{cuDNN and cuBLAS}\label{sec:cudnn}

Much like BLAS or LAPACK are the backbone of HPC computing, cuDNN and cuBLAS form the backbone of the GPU software stacks for DL.
cuDNN is a GPU-accelerated library which provides highly tuned functions that implement DL layers such as convolution, pooling, normalization, activation.
cuBLAS is a GPU-accelerated BLAS library which provides fast implementations of GEMM and GEMV.
The DL layers supported by each API are listed in Table~\ref{tab:cudnnapi}.
And, while there is a wide array of DL frameworks, common between them is the reliance on the primitives defined by cuDNN and cuBLAS.
In fact, all major DL frameworks, such as MXNet, PyTorch, ONNX Runtime, and TensorFlow, rely on cuDNN/cuBLAS API functions for the implementation of common layers.

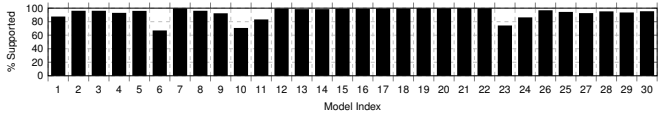
\begin{figure}
\centering
\begin{scaletikzpicturetowidth}{0.48\textwidth}
\begin{tikzpicture}[scale=\tikzscale]
\begin{axis}[
	enlargelimits=0.0,
    ymajorgrids=true,
	x tick label style={
		/pgf/number format/1000 sep=},
    ymajorgrids=true,
    grid style=dashed,
    legend pos=north west,
    legend columns=2,
	ybar interval=0.7,
	ylabel=$\%$ Supported,
	xlabel=Model Index,
	width=0.9\textwidth,
	height=0.1\textwidth,
	cycle list name=Set1,
	cycle list shift=1,
	every axis plot/.append style={fill,draw=none,no markers},
	xticklabels ={1, 2, 3, 4, 5, 6, 7, 8, 9, 10, 11, 12, 13, 14, 15, 16, 17, 18, 19, 20, 21, 22, 23, 24, 26, 25, 27, 28, 29, 30, 31,DUMMY},
    xtick=data,
    bar shift=0pt,
    bar width = 10pt,
    ymin=0,
    legend image code/.code={%
        \draw[#1, draw=none] (0cm,-0.1cm) rectangle (0.2cm,0.1cm);
    },  
]
	\addplot[black!20!black,fill=black!80!black] table [x index=0,y index=1,col sep=comma] {percentage_of_cudnn_supported_layers.dat};

\end{axis}
\end{tikzpicture} 
\end{scaletikzpicturetowidth}
\caption{The percentage of layers supported by cuDNN and cuBLAS (also covered by \dnnscope{})  for each model in Table~\ref{tab:models}.
% {\color{red} Each layer counts as one (not weighted) in calculating the percentage}.
}
 \label{fig:percentage_of_supported_layers}
\end{figure}

Figure~\ref{fig:percentage_of_supported_layers} shows the percentage of layers supported by cuDNN and cuBLAS for each model in Table~\ref{tab:models}.
Most layers within DL models are covered by the cuDNN and cuBLAS API.
The layers that are not supported are non-compute operators (such as concatenate, which joins two tensors across a specified axis) or datatype manipulations (such as reshape, which changes the dimensions of a tensor).
For example, the  cuDNN and cuBLAS functions support $70\%$  of the \texttt{Inception-v2} (ID = $10$) layers.
This is because \texttt{Inception-v2} makes heavy use of unsqueeze --- a tensor reshape layer --- and $27\%$ of the layers in  \texttt{Inception-v2} are unsqueeze layers.

Given a specific DL software stack (e.g. framework, cuDNN, cuBLAS, driver, and other CUDA libraries) and GPU hardware, the cuDNN and cuBLAS functions invoked by a model are fixed. %
Most common layers are supported by cuDNN and cuBLAS and the latency attributed to cuDNN and cuBLAS functions is significant with respect to the model's compute latency.
Figure~\ref{fig:perf_breakdown} shows that for the $30$ vision models, the time spent within the cuDNN and cuBLAS API calls dominates the model's GPU kernel time.
The ``other'' time is either memory operations or framework GPU kernels which are neither cuDNN nor cuBLAS API calls.

Based on the above observations, we propose a ``lower-bound'' latency metric for DL models, which is defined by 
the latencies of the cuDNN and cuBLAS API functions corresponding to the model layers given a specific software/hardware stack.
The ``lower-bound'' latency forms an \textit{ideal} latency, which we use to understand how to improve the model's latency.
We compute the ``lower-bound'' latency under different execution scenarios to determine if optimizations can be made, pinpoint where optimization opportunities are, and quantify the potential benefits of optimizations, as detailed in Section~\ref{sec:design}.

\section{\dnnscope Design and Implementation}\label{sec:design}

\dnnscope consists of four main components: Model Processor, Automatic Benchmark Generator, Performance Database, and Analyzer.
The components are shown in Figure~\ref{fig:dnn_scope_flow} and are used in the benchmarking and analysis workflows:

\begin{itemize}[nosep,leftmargin=0.5em,labelwidth=*,align=left]
\item \textbf{Benchmarking workflow:}
\circledwhite{1} The Model Processor takes ONNX models as input, parses them, performs shape inference, and finds the set of unique layers within the models.
Two layers are considered the same (non-unique) if they have the same operator type, shape, and parameters (i.e. \textbf{only differ in weight values}).
\circledwhite{2} The Automatic Benchmark Generator then generates micro-benchmarks for each unique layer.
The generated micro-benchmarks measure the latency (or the GPU kernel metrics if profiling mode is enabled) of the corresponding cuDNN or cuBLAS function calls for the layers.  %
\circledwhite{3} The micro-benchmarks are then run 
on systems of interest and the results are stored in the Performance Database.

\item \textbf{Analysis workflow:} 
\circledwhitered{4} The user runs the target model using a framework on a system of interest with utilities provided by \dnnscope to get the model execution profile (i.e. the model's latency, cuDNN and cuBLAS logs, and Nsight profile).
\circledwhitered{5} The user then specifies the model and system to \dnnscope.
The model is parsed into layers and the Analyzer queries the latencies of each layer from the Performance Database (using the layers and system information provided) to compute the \question{1} ``lower-bound'' latency under different execution scenarios.
By analyzing the model execution profile and the computed ``lower-bound'', the Analyzer informs optimizations in: \question{2} parallel execution of independent layers, \question{3} convolution algorithm selection, \question{4} framework inefficiency, \question{5} layer fusion, and \question{6} Tensor Core usage.
\end{itemize}

\subsection{\dnnscope Model Processor}\label{sec:processor}

The \circledwhite{1} Model Processor parses ONNX models into \dnnscope's internal %
representation (IR).
The IR wraps around the ONNX Protobuf and has the same layer coverage.
Since ONNX models do not have layer shapes information embedded (except for the input layers), shape inference~\cite{shapeinfer} is performed to determine the shape of each layer.
Layers in the IR (referred to as \textit{layers} and correspond to the ONNX nodes) are annotated with the inferred shapes.
Benchmarks are generated for each layer using its type, shape, and parameters information.

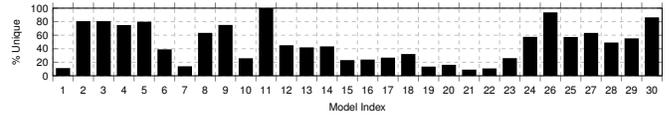
\begin{figure}
\centering
\begin{scaletikzpicturetowidth}{0.48\textwidth}
\begin{tikzpicture}[scale=\tikzscale]
\begin{axis}[
	enlargelimits=0.0,
    ymajorgrids=true,
	x tick label style={
		/pgf/number format/1000 sep=},
    grid style=dashed,
    legend pos=north west,
    legend columns=2,
	ybar interval=0.7,
	ylabel=$\%$ Unique,
	xlabel=Model Index,
	width=0.9\textwidth,
	height=0.1\textwidth,
	cycle list name=Set1,
	cycle list shift=1,
	every axis plot/.append style={fill,draw=none,no markers},
	xticklabels ={1, 2, 3, 4, 5, 6, 7, 8, 9, 10, 11, 12, 13, 14, 15, 16, 17, 18, 19, 20, 21, 22, 23, 24, 26, 25, 27, 28, 29, 30, 31,DUMMY},
    xtick=data,
    bar shift=0pt,
    bar width = 10pt,
    ymin=0,
    legend image code/.code={%
        \draw[#1, draw=none] (0cm,-0.1cm) rectangle (0.2cm,0.1cm);
    },  
]
	\addplot[black!20!black,fill=black!80!black] table [x index=0,y index=1,col sep=comma] {unique_layer_percentage.dat};

\end{axis}
\end{tikzpicture} 
\end{scaletikzpicturetowidth}
\caption{The percentage of unique layers within the $30$ models}
  \label{fig:percentage_of_unique_layers}
\end{figure}

We observe that layers with the same type, shape, and parameters (i.e. \textbf{only differ in weight values}) are repeated extensively within and across models.
Figure~\ref{fig:percentage_of_unique_layers} shows that most models have a low percentage of unique layers --- indicating that layers are repeated extensively within the model.
For example, \texttt{ResNet50-v1} (ID=$17$) has $175$ layers but only $47$ ($26.9\%$) are unique.
The number of unique layers across models of similar architecture is also low.
The \texttt{ResNet*-v1} models (ID=$13,15,17,19,21$) are built from the same  modules and have a total of $1229$ layers, of which only $60$ ($5.6\%$) are unique.
Across all $30$ models, the total number of layers is $5754$, but only $1031$ ($18\%$) are unique.
We exploit this layer repeatability to optimize the benchmark generation and minimize the time to benchmark.
Thus, the Model Processor unifies the repeated layers across the input models and produces a set of unique layers.
The time saved can be used to explore other algorithms and data types (Sections \ref{sec:bengen:algorithm} and \ref{sec:bengen:tensorcore}) benchmarks.

\subsection{Automatic Benchmark Generator}\label{sec:bengen}

The \circledwhite{2} Automatic Benchmark Generator uses the set of unique layers (produced by the Model Processor) and generates C++ code to invoke the benchmark runtime using each layer's type, shape, and parameters information.

\subsubsection{The Benchmark Runtime}\label{sec:design:runtime}

\dnnscope provides a benchmark runtime that measures the latency of the cuDNN or cuBLAS APIs required to execute each layer (as shown in Table~\ref{tab:cudnnapi}).
The runtime also sets up the function arguments for each API.
The setup time is not included in the latency measurement.
The runtime uses the Google Benchmark~\cite{googlebenchmark} library --- a micro-benchmarking support library.
The Google Benchmark library dynamically determines the number of iterations to run each benchmark and ensures that the reported latency results are statistically stable.
Generated benchmarks are linked with the cuDNN/cuBLAS libraries, and are run on systems of interest.

\subsubsection{Algorithm Instantiation}\label{sec:bengen:algorithm}

The convolution layers map to the \texttt{cudnn\allowbreak ConvolutionForward} API (Table~\ref{tab:cudnnapi}).
The convolution API takes one of the following $8$ algorithms as an argument: Implicit GEMM (IGEMM), Implicit PreComputed GEMM (IPGEMM), GEMM, Direct (DRCT), FFT, Tiled FFT (TFFT), Winograd (WING), and Winograd Non-Fused (WINGNF).
These algorithms have different compute and memory characteristics~\cite{anderson2018optimal,ben2018demystifying}.
The optimal algorithm to use depends on the system, layer shape, and layer parameters (e.g. filter size, stride, dilation, etc.)~\cite{cudnn}.
For inference, most frameworks (e.g. MXNet, PyTorch, TensorFlow) rely on the cuDNN provided heuristic function (\texttt{cudnn\allowbreak GetConvolution\allowbreak ForwardAlgorithm}) to choose the convolution algorithm.
The heuristic function suggests an algorithm given the layer's shape, parameters, data type, system, etc.
To explore the design space of algorithm selection, by default, for each layer \dnnscope generates benchmarks using all algorithms applicable to the layer.

\subsubsection{Data Type Support}\label{sec:bengen:tensorcore}

\dnnscope can be configured to generate micro-benchmarks that target different data types.
Both \texttt{float16} and \texttt{float32} are generated by default, but benchmarks can be instantiated for other data types.
The \texttt{float16} benchmarks use Tensor Cores when the API function (see Table~\ref{tab:cudnnapi}) and the system (see Table~\ref{tab:systems}) support it.

\subsubsection{Layer Fusion Support}\label{sec:bengen:fusion}

\dnnscope can be configured to generate micro-benchmarks that target the cuDNN fused API (\texttt{cudnn\allowbreak Convolution\allowbreak BiasActivation\allowbreak Forward}) to perform the convolution, bias, and activation layer sequence.
Two fusion pattern rules are currently handled by \dnnscope: Conv$\rightarrow$Bias$\rightarrow$Activation and Conv$\rightarrow$Bias.
The Conv$\rightarrow$Bias$\rightarrow$Activation maps directly to the fused API.
Fusing Conv$\rightarrow$Bias is implemented through the fused API using  \texttt{CUDNN\_\allowbreak ACTIVATION\_\allowbreak IDENTITY} as the activation function and requires cuDNN version $\geq7.1$.
For older cuDNN versions, the Conv$\rightarrow$Bias is implemented as two calls --- a \texttt{cudnn\allowbreak Convolution\allowbreak Forward} followed by a \texttt{cudnn\allowbreak AddTensor}.
Users can extend \dnnscope's fusion support by registering new fusion patterns as the cuDNN fused API evolves.

\subsubsection{Integration with CUPTI}

\dnnscope can be configured to generate benchmarks that integrate with low-level GPU profiler libraries such as NVIDIA's CUPTI~\cite{cupti}.
This integration allows \dnnscope to capture detailed GPU metrics~\cite{gpumetrics} of benchmarks such as flops, memory transfers, etc.
In this mode, the user specifies the metrics of interest, the number of benchmark iterations for warm-up, and the number of iterations to measure.
\dnnscope does not use the Google Benchmark in this mode since 
a fixed, small number of profiling runs suffice for statistically stable measurement of the metrics.
The profiling outputs (name, timing, and metric values of GPU kernels) 
are stored as metadata to the corresponding benchmark entry in the Performance Database.

\subsection{Performance Database}\label{sec:database}

The \circledwhite{3} benchmarking results are collected and published to \dnnscope{}'s Performance Database.
Each entry within the database is indexed by the system, data type, and layer (type, shape, and parameter information).
The Analyzer queries the database to get the benchmark latencies.
If a query is a miss, then a warning with the information about the missing benchmark is issued to the user and the user is asked if they wish the Automatic Benchmark Generator to generate the missing benchmarks.

\begin{figure}
	\centering
	\includegraphics[trim=0 41 0 50,clip,width=0.35\textwidth]{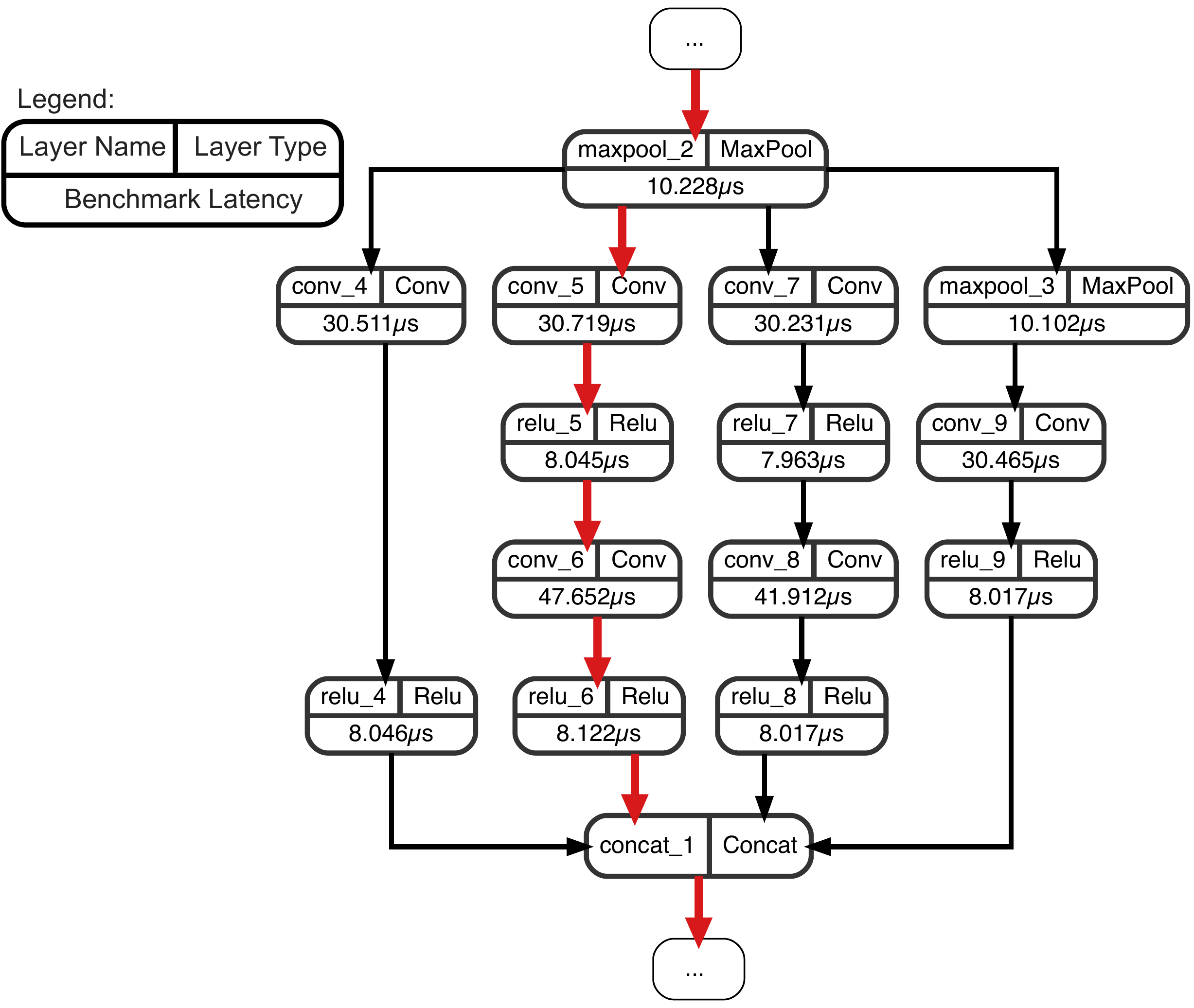}
	\caption{
		The first parallel module of \texttt{Inception-v1} in Figure~\ref{fig:Tesla_V100-SXM2-16GB_normalized} visualized by the \dnnscope{} Analyzer.
		The layers are annotated with the name, type, and latency used for the ``lower-bound'' calculation.
		The critical path used in the parallel mode is highlighted in red.
	}
	\label{fig:inception_v1_module}
\end{figure}

\subsection{\dnnscope Analyzer}\label{sec:analyzer}

The \circledwhitered{4} user runs the target model using a framework on a system of interest with utilities provided by \dnnscope to get the \textit{model execution profile}.
The model execution profile contains information about the model's latency, cuDNN and cuBLAS logs, and Nsight profile (which contains cuDNN/cuBLAS API calls and function backtrace information).
Capturing the model's latency requires the user to place the provided timing functions within their application code.
To capture the usage of cuDNN and cuBLAS functions within a framework,  \dnnscope launches the user code with the \verb|CUDNN_LOGINFO_DBG| and \verb|CUBLAS_LOGINFO_DBG| environment variables.
These environment variables enable the cuDNN and cuBLAS loggers respectively.
Utilities to run the user code using NVIDIA's Nsight profiler are also provided.
The results from Nsight are parsed and correlated with the cuDNN and cuBLAS logs.

The \circledwhitered{5}  user then inputs the model execution profile along with the ONNX model, system, data type.
The model is parsed by the Model Processor into layers.
Then, the \dnnscope Analyzer queries the Performance Database for the benchmark latencies of each layer using the user-specified system and data type (by default \texttt{float32}).
Due to algorithm (Section~\ref{sec:bengen:algorithm}) instantiation, multiple benchmarks may exist for a layer.
The Analyzer, therefore, selects the benchmark result achieving the lowest latency.
The following analyses are then performed:

\subsubsection{\questionbox{\color{white} \small $\mathcal{Q}$\normalfont\sffamily\textsf{1,2}} Sequential and Parallel ``Lower-Bound'' Latency}\label{sec:analyzer:parallel}

DL models may contain layer sequences which can be executed independently in parallel.
The sub-graph formed by these data-independent layer sequences is called a \textit{parallel module}.
For example, a parallel module in \texttt{Inception-v1} is shown in Figure~\ref{fig:inception_v1_module}.
A framework may execute the independent paths within the parallel module either sequentially or in parallel.
Thus, the Analyzer computes the ``lower-bound'' latency of a model using two execution modes: sequential and parallel.

The sequential mode assumes that independent layers are executed sequentially, and therefore is defined as the sum of each layer's benchmark latency.
The parallel strategy assumes that data-independent layers are executed in parallel.
Therefore, the parallel ``lower-bound'' latency is defined by the model's \textit{critical path} --- the simple path from the start to the end layer with the highest latency.
Finding the critical path of a graph is a longest path problem and is NP-hard.
Since a DL model forms a directed acyclic graph (DAG), the critical path can be framed as a shortest path problem~\cite{Sedgewick:2011:ALG:2011916}.
To compute the critical path we construct a weighted DAG from the model graph where the edge weight between two nodes (layers) is negative of the latency of the layer at the tail of the edge.
Computing the shortest path from the start to the end layer of the constructed weighted DAG produces the critical path of the model.
The parallel ``lower-bound'' latency is the sum of layers latencies along the critical path.
\dnnscope{} visualizes the critical path of the model (e.g. Figure~\ref{fig:inception_v1_module}), and 
the difference between the sequential and parallel ``lower-bound'' latencies indicates the profit of executing independent layers in parallel.
Other analyses performed by \dnnscope{} leverage the sequential and parallel ``lower-bound'' latencies, and the benefits can be calculated in terms of either sequential or parallel mode.

\subsubsection{\questionbox{\color{white} \small $\mathcal{Q}$\normalfont\sffamily\textsf{3}} Convolution Algorithm Selection}\label{sec:analyzer:algorithm}

The Analyzer uses the parsed cuDNN log in the model execution profile to determine  if the cuDNN algorithm  used by the framework for each layer is optimal (recall from Section~\ref{sec:bengen:algorithm} that benchmark results using  all available algorithms for layers exist in the Performance Database).
Cases where the algorithm choice 
is sub-optimal are reported to the user along with how much latency improvement could be gained if algorithm selection was ideal.
The user can act upon these suggestions by forcing the framework to use a specific algorithm for each layer.

\subsubsection{\questionbox{\color{white} \small $\mathcal{Q}$\normalfont\sffamily\textsf{4}} Framework Inefficiency Inspection}\label{sec:analyzer:framework}

The expected cuDNN and cuBLAS API calls are known to the Analyzer from the ``lower-bound'' latency computation.
The Analyzer compares the model execution profile against the expected execution to pinpoint inefficiencies within the framework.
The user is presented with any deviation observed in cuDNN or cuBLAS API invocation's parameters or their execution order.
CUDA API functions and CUDA kernels executed between cuDNN or cuBLAS API calls, are also presented to the user --- along with their backtraces.

\subsubsection{\questionbox{\color{white} \small $\mathcal{Q}$\normalfont\sffamily\textsf{5}} Layer Fusion Analysis}\label{sec:analyzer:fusion}

If the user enables the benchmark generation for layer fusion (as described in Section~\ref{sec:bengen:fusion}), then the Analyzer can be used to determine the potential profitability if layer fusion is employed.
The Analyzer traverses the model layers and looks for the fusion pattern rules (listed in Section~\ref{sec:bengen:fusion}).
If one of these patterns is found, then the corresponding fused operation's latency is queried from the database and is used in the ``lower-bound'' computation (in either sequential or parallel mode).
If the benchmark is unavailable, or failed to run, then the latencies of the non-fused layers are used.
The difference between the non-fused ``lower-bound'' latency and the fused ``lower-bound'' latency determines the profitability of layer fusion.

\subsubsection{\questionbox{\color{white} \small $\mathcal{Q}$\normalfont\sffamily\textsf{6}} Tensor Core Analysis}\label{sec:analyzer:tensorcore}

The Analyzer determines if the target model execution utilizes Tensor Cores by looking at kernel names in the model execution profile.
Kernel names that match the \verb!_[ish]\d+*! Regular-expression use Tensor Cores.
By default, benchmarks targeting both \texttt{float16} and \texttt{float32} are generated.
When benchmarks are run on systems with Tensor Core support, the difference between the ``lower-bound'' latency of \texttt{float32} and \texttt{float16} informs the profitability of using Tensor Cores with \texttt{float16}.

\subsection{Sustainability and Extensibility}\label{sec:extensibility}

The sustainability of \dnnscope{} is ensured by providing an automated benchmark generation and analysis workflow design along with a continuously updated Performance Database.
Benchmarking requires limited effort, as the micro-benchmarks are automatically generated, and the user only needs to compile and run the generated code on systems of interest.
% The Performance Database is continuously updated with new benchmark results.
A big insight of the proposed design is that there is ample layer repeatability within and across models.
This keeps the number of unique layers and thus the number of Performance Database entries in check over time.
For new models, only the newly introduced unique layers are benchmarked.

For example, consider a scenario where all models in Table~\ref{tab:models} except for \texttt{ResNet*-v2} have already been benchmarked and the results are in the Performance Database.
Using our design, benchmarking the \texttt{ResNet*-v2} models requires measuring all the \texttt{ResNet*-v2}  layers that are not within the Performance Database.
Evaluating this hypothetical scenario results in a $75\%$ reduction ($30$ minutes) in benchmarking time on the \texttt{Tesla\_V100} system for batch size $32$.
The saving would be even larger on slower systems.
By storing and reusing the micro-benchmark results in the Performance Database 
we minimize the time cost of running micro-benchmarks. 

\dnnscope is extensible.
As shown in Figure~\ref{fig:dnn_scope_flow}, \dnnscope is designed as a set of modular components.
As new cuDNN functions are introduced, users update the \dnnscope{} runtime accordingly. 
For example, if a new cuDNN convolution algorithm is added, then the user can just add it to the list of algorithms to instantiate in the convolution benchmark implementation.
If a new cuDNN/cuBLAS API or a fused API is added, then a user needs to add the benchmark implementation for the new API using the templates provided by \dnnscope{}.
Users can also extend the Automatic Benchmark Generator to support other runtimes that target other software libraries or hardware, and leverage most of the other components unmodified.
These runtimes can target the frameworks' Python or C++ API or other DL libraries (e.g. MIOpen~\cite{jeh2019miopen} on AMD GPUs, or MKL-DNN~\cite{mkldnn} on CPUs).
Through the novel benchmarking and analysis design, \dnnscope{} copes well with the fast evolving pace of DL innovations.

\section{Evaluation}\label{sec:eval}

We implemented \dnnscope and evaluated its design by answering \question{1-6}~.
We evaluated $30$ ONNX models (listed in Table~\ref{tab:models}) in the MXNet (v$1.5.1$), ONNX Runtime (v$0.5.0$), and PyTorch (v$1.3$) frameworks.
Experiments were run on the $7$ systems listed in Table~\ref{tab:systems}.
All systems use Ubuntu $18.04.3$ LTS, CUDA $10.1.243$, cuDNN Version $7.6.3$, and CUDA Driver $430.26$.
The micro-benchmarks were compiled with GCC $7.4.0$.
We first computed the \texttt{float32} ``lower-bound'' latency in both sequential and parallel modes.
Then we used the Analyzer to uncover and explore optimization opportunities --- cuDNN heuristics, framework inefficiencies, layer fusion, and usage of Tensor Cores,
and show their
impact on the latency.
% The reader is encouraged to explore further documentation and experimentation at \websiteurl~.

\begin{table*}
	\centering
	\caption{
		We used $7$ GPU systems for evaluation. The systems cover the past GPU generations (from Kepler to the latest Turing).
		Amazon cloud (AWS) is used for $4$ of the systems and the other $3$ are local machines.
		The $4$ Turing and Volta GPUs support Tensor Cores and their theoretical Tensor Core performance (Tensor TFLOPS) are listed.
	}
	\resizebox{0.98\textwidth}{!}{%

		\begin{tabular}{l l l l r r r} \toprule
			\centering%
			\textbf{\thead{Name}} & \textbf{\thead{CPU}} & \textbf{\thead{GPU (Release Year)}} & \textbf{\thead{\shortstack{GPU}\\{Architecture}}} & \textbf{\thead{\shortstack{GPU Memory}\\{Capacity, Bandwidth}}} & \textbf{\thead{\shortstack{\textbf{Theoretical}\\\textbf{FP32 TFLOPS}}}}  & \textbf{\thead{\shortstack{\textbf{Theoretical}\\\textbf{Tensor TFLOPS}}}} \\ \midrule
			Tesla\_K80 (AWS P2)   & Intel Xeon CPU E5-2686 v4      & Tesla K80 (2014)                    & Kepler                                                                  & 12 GB, 480 GB/s                        & 5.6  & \no   \\
			Tesla\_M60 (AWS G3)   & Intel Core i9-7900X CPU        & Tesla M60 (2015)                    & Maxwell                                                                 & 7 GB, 160.4 GB/s                       & 4.8  & \no   \\
			TITAN\_ Xp            & Intel Xeon CPU E5-2686 v4      & TITAN Xp (2017)                     & Pascal                                                                  & 12 GB, 547.6 GB/s                      & 12.2 & \no   \\
			TITAN\_V              & Intel Core i7-7820X CPU        & TITAN V (2017)                      & Volta                                                                   & 12 GB, 672 GB/s                        & 14.9 & 110.0 \\
			Tesla\_V100 (AWS P3)  & Intel Xeon CPU E5-2686 v4      & Tesla V100 SXM2 (2018)              & Volta                                                                   & 16 GB, 900 GB/s                        & 15.7 & 125.0 \\
			Quadro\_RTX           & Intel Xeon CPU E5-2630 v4      & Quadro RTX 6000 (2019)              & Turing                                                                  & 24 GB, 624 GB/s                        & 16.3 & 130.5 \\
			Tesla\_T4 (AWS G4)    & Intel Xeon Platinum 8259CL CPU & Tesla T4 (2019)                     & Turing                                                                  & 15 GB, 320 GB/s                        & 8.1  & 65.0  \\
			\bottomrule
		\end{tabular}%
	}
	\label{tab:systems}
\end{table*}

\subsection{``Lower-Bound'' Latency vs. Measured Latency}\label{sec:correlation}

\begin{figure}
	\centering
	\includegraphics[width=0.48\textwidth]{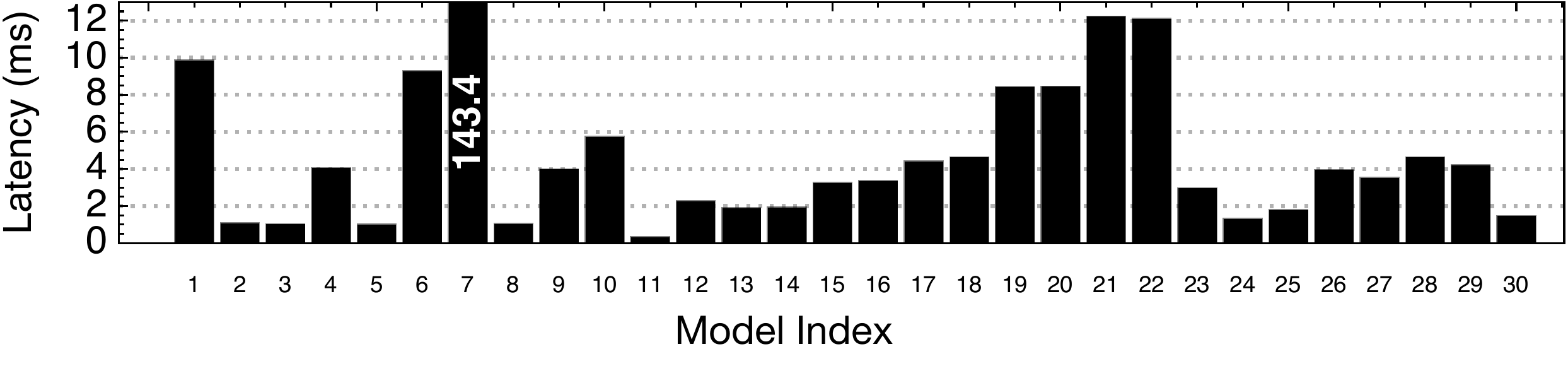}
	\caption{
		The measured latency of all ONNX models using batch size 1 with MXNet backend on Tesla\_V100 in Table~\ref{tab:systems}.
	}
	\label{fig:mxnet_latency_v100}
\end{figure}
\begin{figure}
	\centering
	\includegraphics[width=0.48\textwidth]{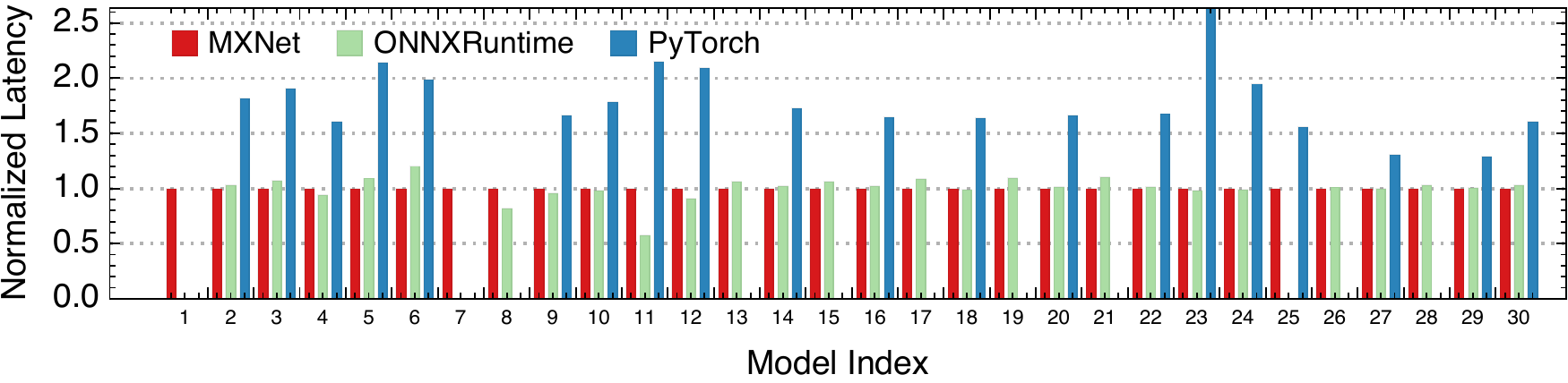}
	\caption{
		The measured latency of all ONNX models with MXNet, ONNX Runtime, and PyTorch backends (normalized to MXNet latency) using batch size 1 on Tesla\_V100.
	}
	\label{fig:framework_compare}
\end{figure}

\begin{figure}
	\centering
	\includegraphics[width=0.48\textwidth]{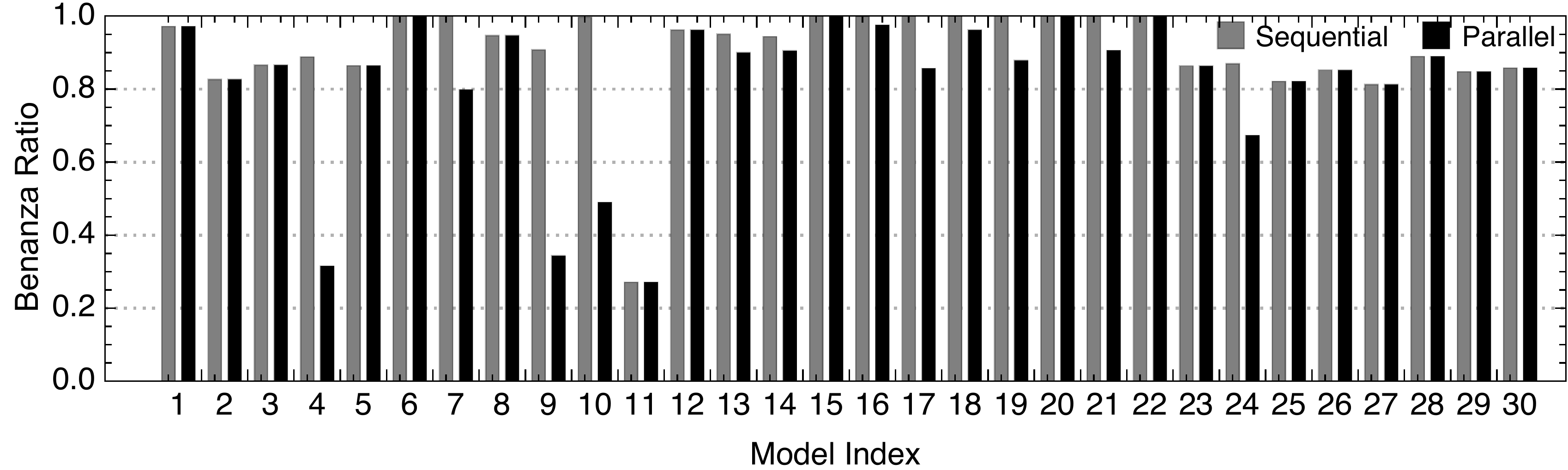}
	\caption{
		The \dnnscope{} Ratio in sequential and parallel mode of 30 models in MXNet using batch size 1 on \texttt{Tesla\_V100}.
	}
	\label{fig:Tesla_V100-SXM2-16GB_normalized}
\end{figure}
\begin{figure}
  \centering
	\includegraphics[width=0.48\textwidth]{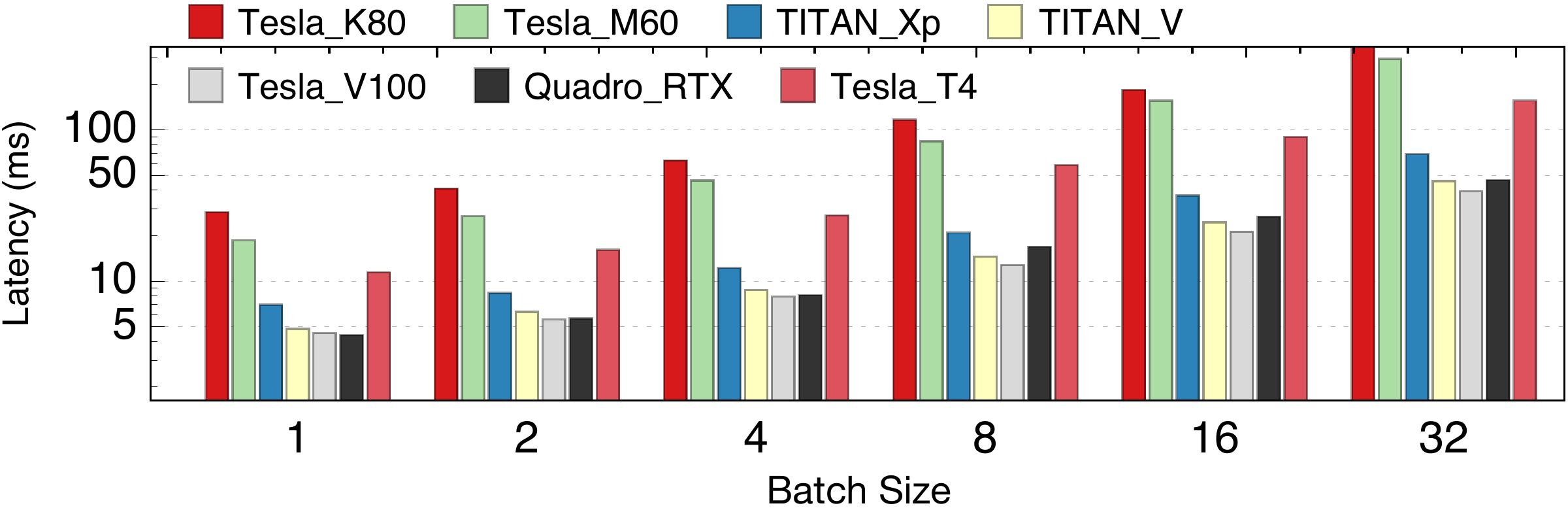}
	\caption{
		The measured latency of \texttt{ResNet50\_v1} in MXNet across batch sizes and systems.
	}
	\label{fig:ResNet050-v1_latency}
\end{figure}

We measured the inference latency of the $30$ models using MXNet, ONNX Runtime, and PyTorch on the \texttt{Tesla\_V100} system.
Figure~\ref{fig:mxnet_latency_v100} shows the measured latency across all models and Figure~\ref{fig:framework_compare} compares the latencies using different frameworks.
Due to the lack of support of some ONNX operators by ONNX Runtime~\cite{onnxruntime} and PyTorch~\cite{paszke2017pytorch}, not all models run within these frameworks.
As MXNet is the fastest in general, subsequent sections of the paper (with the exception of Section~\ref{sec:framework})
focus on informing optimizations in MXNet.

\subsubsection{\questionbox{\color{white} \small $\mathcal{Q}$\normalfont\sffamily\textsf{1,2}} Sequential Mode vs Parallel Mode}\label{sec:seq_par}

The difference between the ``lower-bound'' latency and the measured latency indicates the optimization opportunities in the framework and its use of the cuDNN and cuBLAS APIs.
A model's ``lower-bound'' latency normalized to its measured latency is referred to as its \textit{\dnnscope Ratio} (BR).
Figure~\ref{fig:Tesla_V100-SXM2-16GB_normalized} shows the BR in sequential (\textit{BR\textsubscript{sequential}}) and parallel mode (\textit{BR\textsubscript{parallel}}) in MXNet across all models using batch size 1 on the \texttt{Tesla\_V100} system.

The BR\textsubscript{sequential} across models has a geometric mean of $0.88$, thus a potential latency speedup of $\frac{1.0}{0.88} = 1.14\times$ can be made to the measured model execution.
The BR\textsubscript{parallel} across models has a geometric mean of $0.76$, indicating a potential latency speedup of $\frac{1.0}{0.76} = 1.32\times$.
The difference between a model's parallel and sequential ``lower-bound'' latency depends on the existence of parallel modules within the model and how compute-intensive the data-independent paths are.
Models without parallel modules have the same sequential and parallel ``lower-bound'' latency, thus the BR\textsubscript{sequential} is equal to the BR\textsubscript{parallel}.
For models with compute-intensive parallel modules, such as the Inception models (ID=$4,9,10$), the potential speedup of the latency (or $\frac{1}{\text{BR\textsubscript{parallel}}}$) is $2.87\times$, $2.69\times$, and $2.45\times$ respectively.
The BR\textsubscript{sequential} and BR\textsubscript{parallel} of \texttt{LeNet} (ID=$11$) are both low because \texttt{LeNet} is a simple model which has low latency ($0.33ms$ as shown in Figure~\ref{fig:mxnet_latency_v100}) and the MXNet overhead and other non-compute portion is high, thus its BR is low.

The sequential ``lower-bound'' latency of the models with parallel modules (e.g. \texttt{Inception} and \texttt{ResNet} models)  is closer to their measured latency when compared to the parallel ``lower-bound'' latency (BR\textsubscript{parallel} $<$ BR\textsubscript{sequential} $< 1$).
This suggests that parallel modules are executed sequentially in MXNet, even though the data-independent layers could be run in parallel.
We verified the sequential execution behavior in MXNet by inspecting the model execution profile.
Thus we evaluated the benefits of the latter optimizations in terms of the sequential ``lower-bound'' latency.

\begin{figure}
  \centering
	\includegraphics[width=0.48\textwidth]{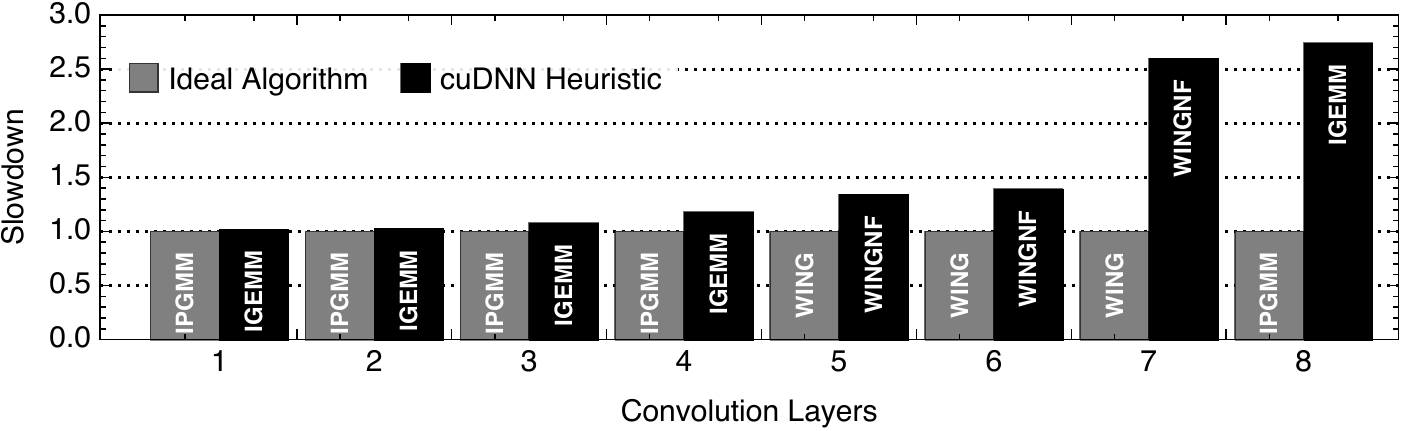}
	\caption{
	The cuDNN heuristic selects $8$ non-optimal convolution layer algorithms for \texttt{ResNet50\_v1} using batch size $32$ on \texttt{Tesla\_V100}.
	Up to $2.75\times$ speedup can be achieved if selection was ideal.
	}
	\label{fig:cudnn_herusitics}
\end{figure}
\begin{figure}
  \centering
	\includegraphics[width=0.48\textwidth]{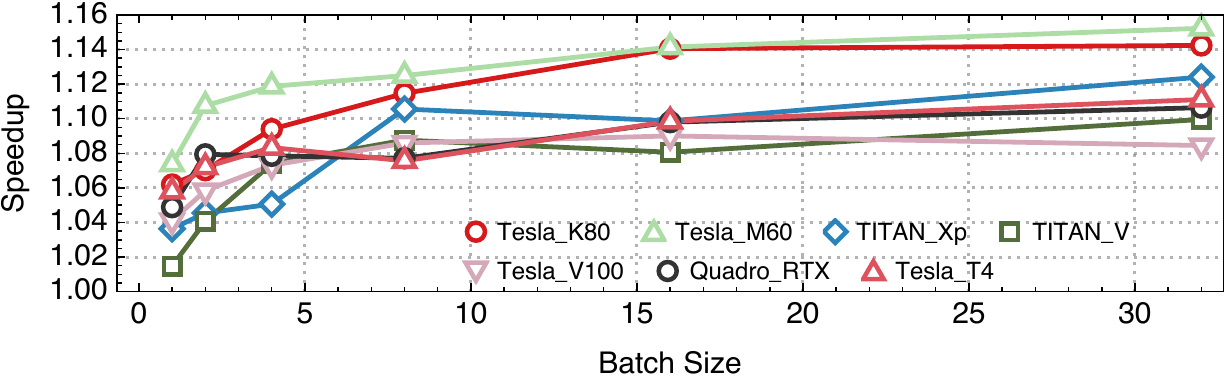}
	\caption{The latency speedup achieved for \texttt{ResNet50\_v1} by applying the MXNet optimization described in Section~\ref{sec:evaluation:mxnet}
	across batch sizes and systems.
	}
	\label{fig:mxnet_opt}
\end{figure}

\begin{figure*}
	\centering
	\begin{minipage}[t]{0.32\linewidth}
		\includegraphics[trim=0 34 0 0,clip,width=0.98\textwidth]{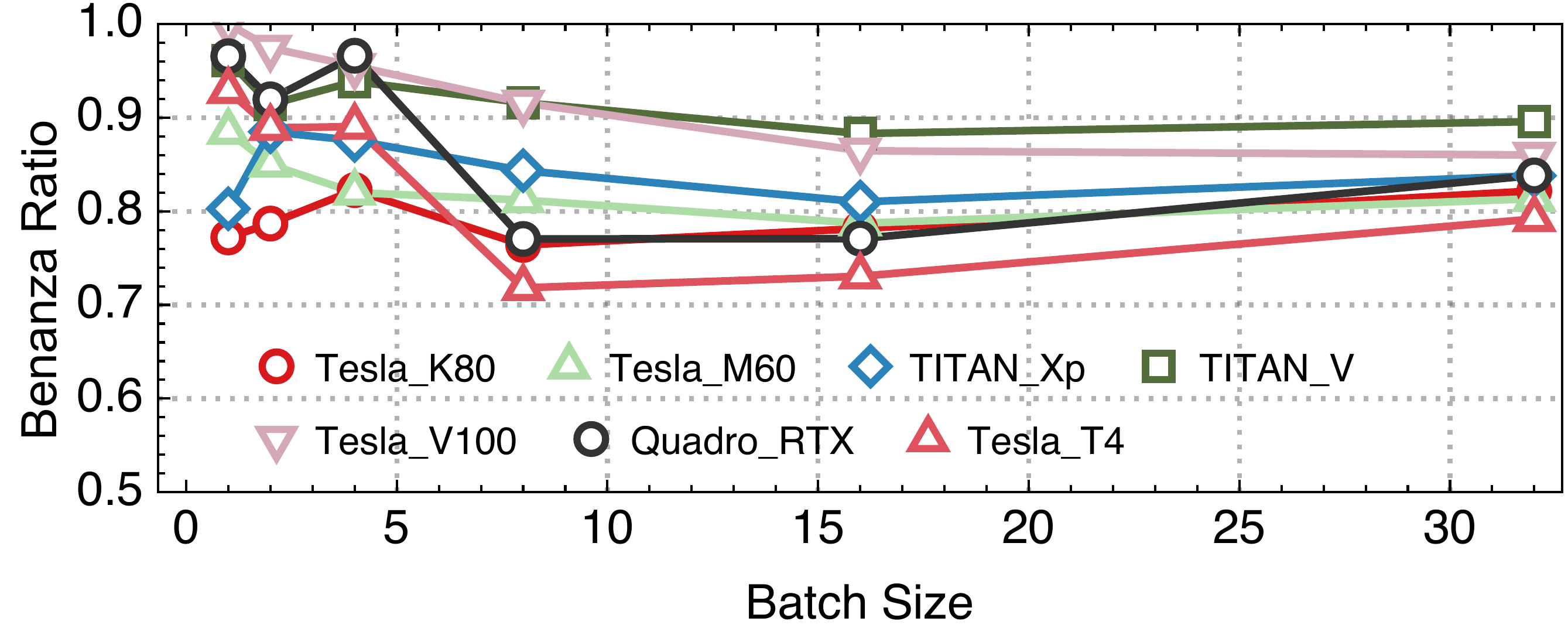}
		\caption{
		    The BR\textsubscript{sequential} of \texttt{ResNet50-v1}.
		}
		\label{fig:resnet50_v1_serial_geomean_mxnet}
	\end{minipage}%
	\hfill%
	\begin{minipage}[t]{0.32\linewidth}
		\centering
		\includegraphics[trim=0 28 0 0,clip,width=0.98\textwidth]{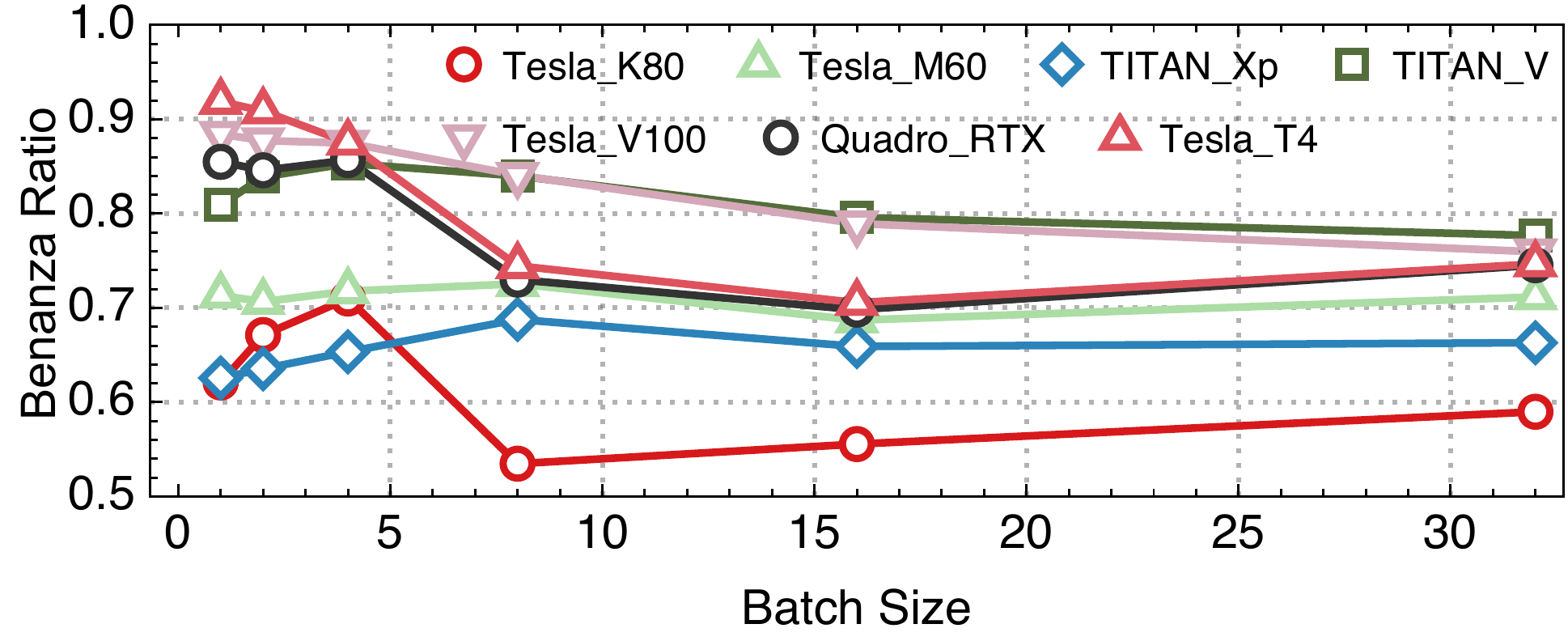}
		\caption{
			The geometric mean of the BR\textsubscript{sequential} of all models.
		}
		\label{fig:serial_geomean_mxnet}
	\end{minipage}
	\hfill%
	\begin{minipage}[t]{0.32\linewidth}
		\centering
		\includegraphics[trim=0 18 0 0,clip,width=0.98\textwidth]{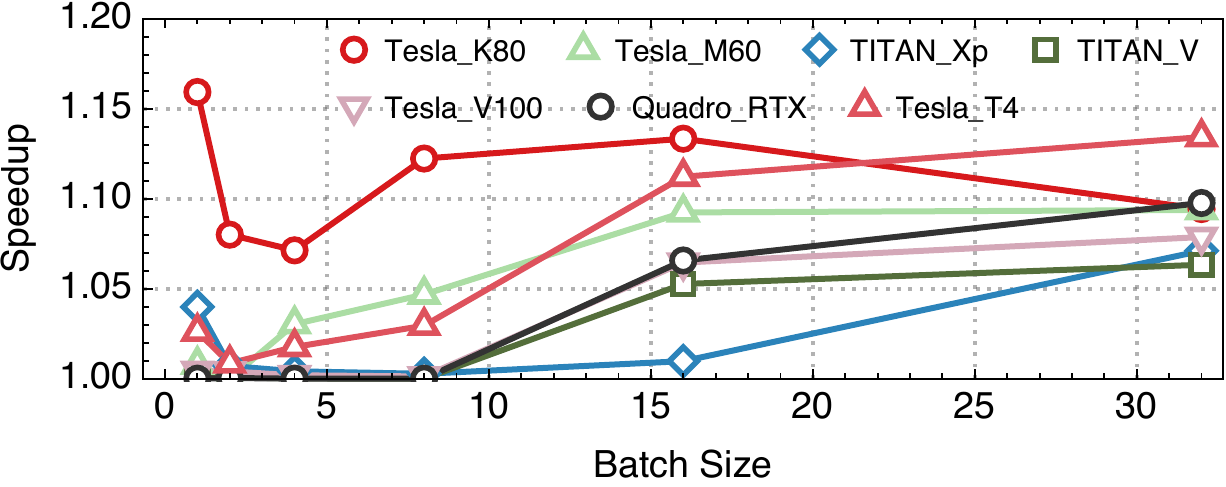}
		\caption{
		    The latency speedup for \texttt{ResNet50-v1} if the cuDNN heuristic selections were optimal.
		}
		\label{fig:end_to_end_latency_improvement}
	\end{minipage}%
	\hfill%
	\begin{minipage}[t]{0.32\linewidth}
		\centering
		\includegraphics[trim=0 18 0 0,clip,width=0.98\textwidth]{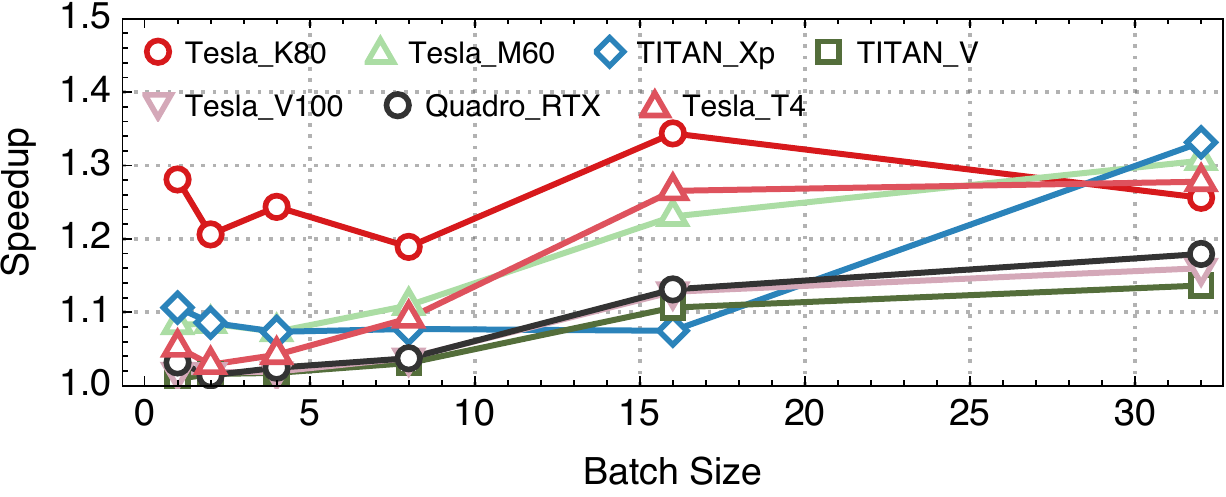}
		\caption{
			The geometric mean of the latency speedup for all models  by using the optimal convolution algorithm. %
		}
		\label{fig:end_to_end_latency_geomean_improvement}
	\end{minipage}%
	\hfill%
	\begin{minipage}[t]{0.32\linewidth}
		\centering
		\includegraphics[trim=0 18 0 0,clip,width=0.98\textwidth]{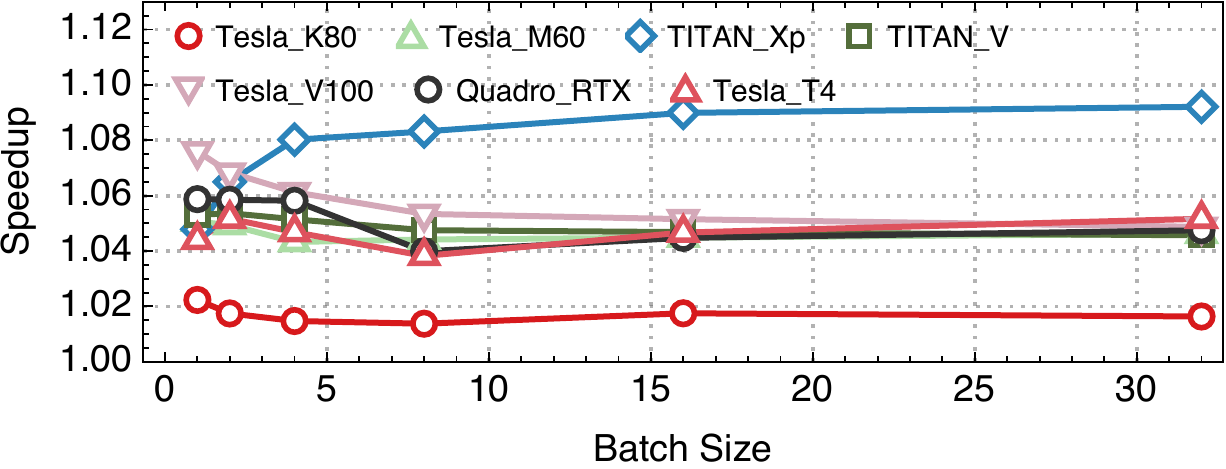}
		\caption{
			The latency speedup for \texttt{ResNet50-v1} if layer fusion was performed.
		}
		\label{fig:end_to_end_fused_latency_improvement}
	\end{minipage}%
	\hfill%
	\begin{minipage}[t]{0.32\linewidth}
		\centering
		\includegraphics[trim=0 18 0 0,clip,width=0.98\textwidth]{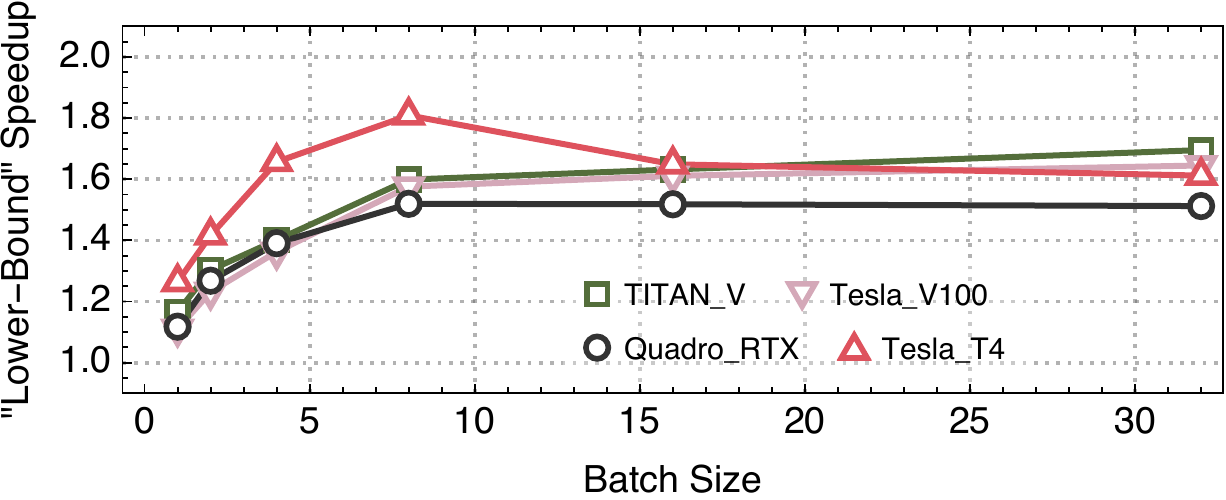}
		\caption{
			The ``lower-bound'' latency speedup if Tensor Cores (\texttt{NCHW}) were used for \texttt{ResNet50-v1}.
		}
		\label{fig:end_to_end_float32_nchw_latency_improvement}
	\end{minipage}%
	\hfill%
	\begin{minipage}[t]{0.32\linewidth}
		\centering
		\includegraphics[trim=0 18 0 0,clip,width=0.98\textwidth]{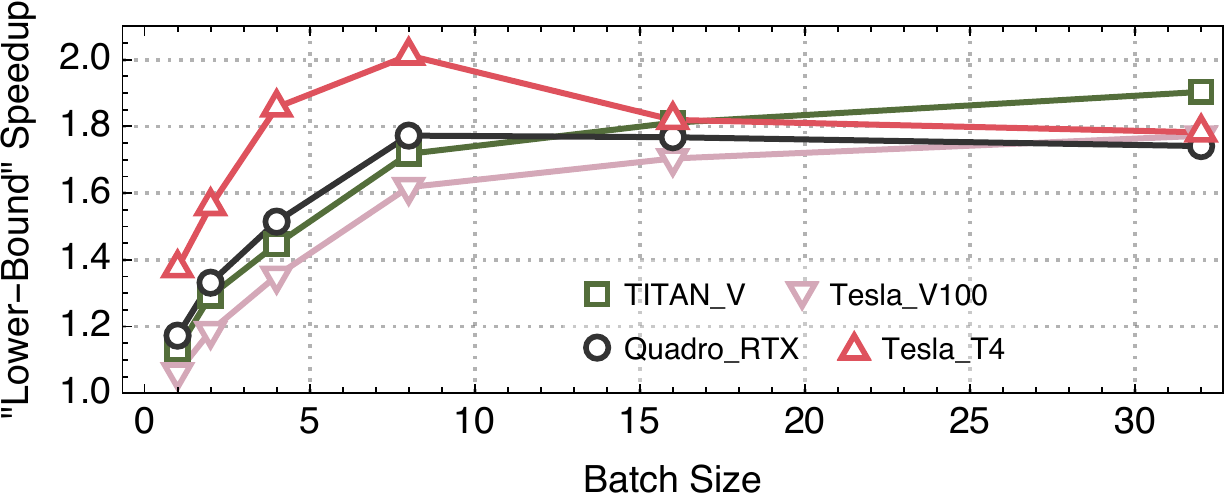}
		\caption{
			The ``lower-bound'' latency speedup for \texttt{ResNet50-v1} if Tensor Cores (\texttt{NHWC}) were used.
		}
		\label{fig:end_to_end_float32_nhwc_latency_improvement}
	\end{minipage}%
	\hfill%
    \begin{minipage}[t]{0.32\linewidth}
    	\centering
    	\includegraphics[trim=0 18 0 0,clip,width=0.98\textwidth]{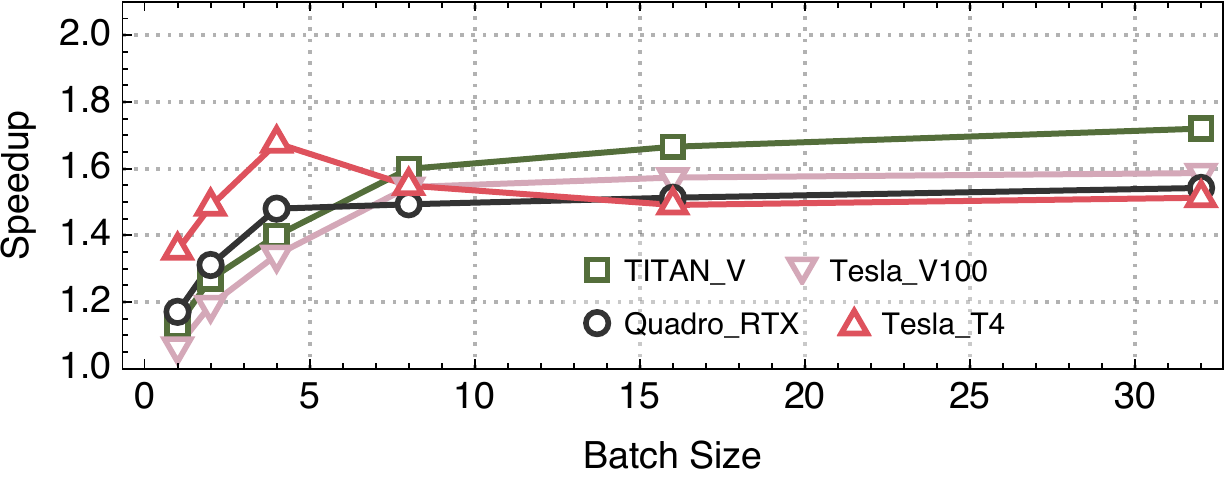}
    	\caption{
    	The latency speedup for \texttt{ResNet50-v1} if Tensor Cores (\texttt{NHWC}) were used.
    	}
    	\label{fig:end_to_end_float32_latency_improvement}
    \end{minipage}%
	\hfill%
    \begin{minipage}[t]{0.32\linewidth}
    	\centering
    	\includegraphics[trim=0 18 0 0,clip,width=0.98\textwidth]{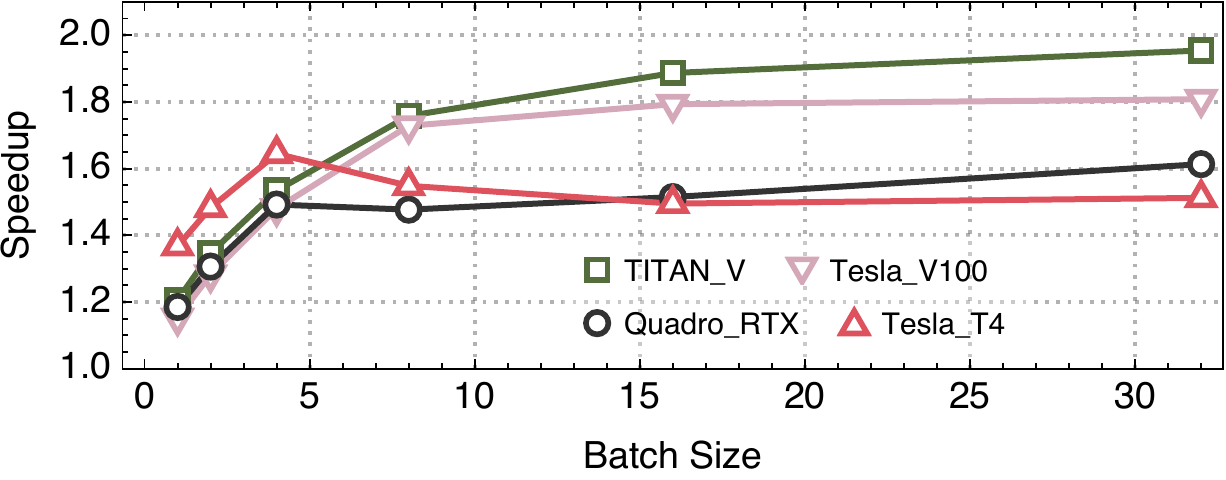}
    	\caption{
    	The latency speedup for \texttt{ResNet50-v1} if parallel execution, optimal algorithm selections, layer fusion, and Tensor Cores (\texttt{NHWC}) were used.
    	}
    	\label{fig:end_to_end_fused_nhwc_float32_latency_improvement}
    \end{minipage}
\end{figure*}

\subsubsection{Batch Sizes and Systems}\label{sec:batch_machine}

To demonstrate \dnnscope{}'s functions across batch sizes and systems, we evaluated the ``lower-bound'' latency of all models using different batch sizes from $1$ to $32$ on representative systems (shown in Table~\ref{tab:systems}).
We select batch size $32$, since some models cannot be run using batch sizes beyond $32$ due to GPU memory limitations.
Figure~\ref{fig:ResNet050-v1_latency} shows the measured latency of \texttt{ResNet50-v1} on all systems in log scale.
As expected, latencies are reversely correlated to the compute capability of the system (e.g. theoretical FP32 TFLOPS in Table~\ref{tab:systems}).
\texttt{ResNet50-v1} has a higher latency on \texttt{Quadro\_RTX} when compared to \texttt{Tesla\_V100}, since \texttt{Quadro\_RTX} has an on-chip (global) memory bandwidth of $624$ GB/s whereas \texttt{Tesla\_V100} has an on-chip memory bandwidth of $900$ GB/s.

Figure~\ref{fig:resnet50_v1_serial_geomean_mxnet} shows the BR\textsubscript{sequential} of \texttt{ResNet50-v1} across batch sizes and systems.
The results suggest that \texttt{ResNet50-v1}'s optimization opportunities are system and batch size dependent.
Both \texttt{Tesla\_V100} and \texttt{TITAN\_V} are highly optimized to run \texttt{ResNet50-v1} across batch sizes, since their BR is high --- ranging from $0.86$ to $1.0$.
The BR for \texttt{Tesla\_T4} and \texttt{Quaro\_RTX} is high for batch sizes $1$ to $4$ but drops beyond that.
\texttt{ResNet50-v1} is less optimized on the other systems
and has a low BR.

The geometric mean of the BR\textsubscript{sequential} for all the models across systems and batch sizes is shown in Figure~\ref{fig:serial_geomean_mxnet}.
Both \texttt{Tesla\_V100} and \texttt{TITAN\_V} still have a high BR ($0.76-0.88$).
A drop was still observed for \texttt{Tesla\_T4} and \texttt{Quaro\_RTX} at batch size $4$.
\texttt{Tesla\_M60} and \texttt{TITAN\_Xp} have a BR between $0.63$ and $0.72$.
The oldest GPU generation, \texttt{Tesla\_K80}, has the lowest BR and is the least optimized.

Overall, the current software stack (latest MXNet, cuDNN, and CUDA libraries used in the evaluation) is more optimized for the recent GPU generations (Turing and Volta) using smaller batch sizes.
Compared to Volta, the software stack is less optimized for Turing.
This is possibly because Turing is newly released, and we expect optimizations that target Turing to increase.
Moreover, the low BR for the older GPUs suggest that vendors prioritize optimizations for newer GPU generations over older ones.

\begin{figure}
	\centering
	\includegraphics[width=0.48\textwidth]{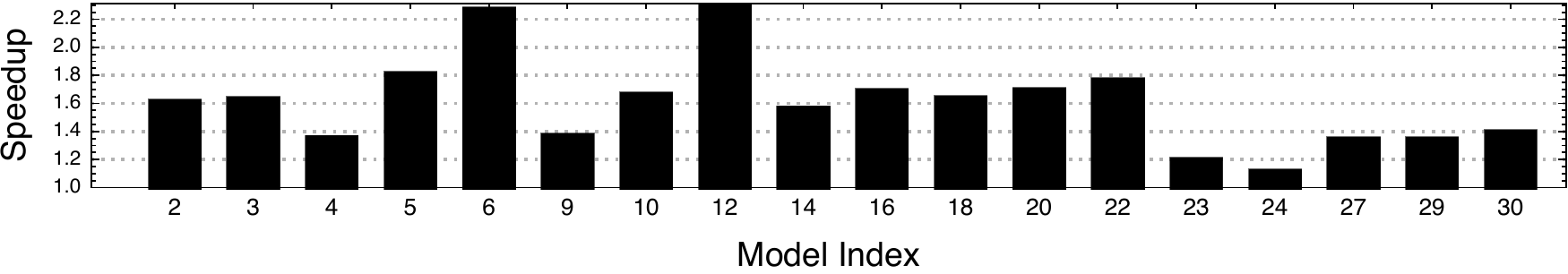}
	\caption{The speedup achieved by removing unnecessary cuDNN API synchronizations in PyTorch on \texttt{Tesla\_V100} using batch size 1.
	}
	\label{fig:caffe2_opt}
\end{figure}

\subsection{\questionbox{\color{white} \small $\mathcal{Q}$\normalfont\sffamily\textsf{3}} cuDNN Convolution Heuristics}\label{sec:heuristics}

Using the \dnnscope Analyzer, we observed that heuristics employed by cuDNN (and subsequently the frameworks) are not always optimal.
For example, Figure~\ref{fig:cudnn_herusitics} shows the convolution layer latencies using the algorithms informed by cuDNN heuristics  (labeled as \textit{cuDNN Heuristic}) normalized to using the optimal algorithm (labeled as \textit{Ideal Algorithm}) for \texttt{ResNet50\_v1} using batch size 32 on \texttt{Tesla\_V100}. 
The algorithm choices are listed in Section~\ref{sec:bengen:algorithm}.
Figure~\ref{fig:end_to_end_latency_improvement} shows the latency speedup for \texttt{ResNet50\_v1} across batch sizes and systems  by using the optimal convolution algorithm for all convolution layers.
Figure~\ref{fig:end_to_end_latency_geomean_improvement} shows the geometric mean of the latency speedup for all models by using the optimal algorithms.
At batch size $32$, the speedup ranges between $1.14\times$ and $1.32\times$ across GPUs.
Both the latest and older GPU architectures can benefit from better algorithm heuristics.

\subsection{\questionbox{\color{white} \small $\mathcal{Q}$\normalfont\sffamily\textsf{4}} Inefficiencies in Frameworks}\label{sec:framework}

We used \dnnscope to identify the inefficiencies in MXNet and PyTorch.
We then implemented the optimizations informed by \dnnscope and show the latency speedup after the framework modifications.

\subsubsection{MXNet ONNX Model Loader}\label{sec:evaluation:mxnet}

We observed through the Analyzer that there are layers in the model execution profile where the cuDNN API arguments deviate from what is expected.
An inspection of the Analyzer's parsed Nsight profile pointed to an \verb|image_2d_pad_constant_kernel| GPU kernel function being invoked before every convolutional layer.
Non-zero padding leads to the observed deviation between the expected and actual cuDNN API calls.
We inspected the MXNet source code and found that padding layers are inserted during the loading of ONNX models in MXNet.
ONNX supports specifying asymmetric padding as a parameter in convolution layers, whereas MXNet does not.
Therefore, MXNet must insert padding layers before convolution layers where asymmetric padding is used when loading ONNX models.
However, the MXNet ONNX model loader adds padding layers before every convolution layer (regardless of  the use of asymmetric padding).  %
A non-intrusive optimization is to only insert padding layers if asymmetric padding is used.
With this simple one-line optimization, we observed up to $1.15\times$ latency speedup for  \texttt{ResNet50-v1} (shown in Figure~\ref{fig:mxnet_opt}).

\subsubsection{PyTorch cuDNN Wrapper}\label{sec:evaluation:caffe2}

Using \dnnscope we observed that there were excessive calls to \texttt{cudaStreamWaitEvent} between cuDNN API calls.
Using the backtrace information from the model execution profile, we identified the PyTorch source file that introduces these synchronizations.
Upon further study of the source code, we found that all cuDNN functions are invoked by a cuDNN wrapper in PyTorch.
The wrapper manages a pool of cuDNN handles and is designed to enable invoking cuDNN functions from different CPU threads.
cuDNN functions managed by the same handle are synchronized and executed sequentially.
In the current PyTorch (v$1.3$), however, a single handle is used for inference, and thus forced synchronization occurs before each cuDNN function call.
The synchronizations cause $100\mu s$ stalls on average between cuDNN functions, thus the latency saved through this optimization is a function of the number of layers in a model.
We modified PyTorch to elide the cuDNN wrapper and only synchronize before and after performing inference.
Figure~\ref{fig:caffe2_opt} shows the speedup achieved by this optimization for batch size 1.
\texttt{MobileNet-v2} (ID=$12$) achieves a $2.3\times$ speedup, since it has low latency and a large number of layers.

\subsection{\questionbox{\color{white} \small $\mathcal{Q}$\normalfont\sffamily\textsf{5}} Layer Fusion}\label{sec:fusion}

We used \dnnscope to evaluate the potential benefits of layer fusion.
Figure~\ref{fig:end_to_end_fused_latency_improvement} shows the latency speedup from layer fusion for \texttt{ResNet50-v1} across the systems.
\texttt{ResNet50-v1}  has the layer sequence pattern Conv$\rightarrow$Bias$\rightarrow$BatchNorm$\rightarrow$Activation.
\dnnscope reports that the Conv$\rightarrow$Bias sequence can be fused for better latency and performs the fusion analysis (Section~\ref{sec:analyzer:fusion}).
In all, $64$ ($18\%$) layers were fused and up to $1.09\times$ speedup was achieved over the measured latency across systems for \texttt{ResNet150-v1}.
By inspecting the model execution profile, we found no indication that MXNet, ONNX Runtime, or PyTorch perform layer fusion using the cuDNN fused API.

\subsection{\questionbox{\color{white} \small $\mathcal{Q}$\normalfont\sffamily\textsf{6}} Tensor Cores}\label{sec:tensorcore}

We used \dnnscope to evaluate the potential benefits of using \texttt{float16} and Tensor Cores available on recent GPU architectures.
While the cuDNN Tensor Core API supports both \texttt{NHWC} and \texttt{NCHW} layout, NVIDIA recommends the use of \texttt{NHWC}.
We use \dnnscope to generate benchmarks targeting both the \texttt{NHWC} and \texttt{NCHW} layout and evaluated the ``lower-bound'' latency speedup, as shown in Figures~\ref{fig:end_to_end_float32_nhwc_latency_improvement} and \ref{fig:end_to_end_float32_nchw_latency_improvement} respectively.
As expected, using the \texttt{NHWC} achieves higher speedup.
Internally, the current cuDNN API implements \texttt{NCHW} convolutions in terms of \texttt{NHWC} with an implicit transposition.
As compute dominates (i.e. larger batch sizes), the relative overhead of the transposition becomes small; hence, \texttt{NCHW} and \texttt{NHWC} have similar performance for larger batch sizes.
Figure~\ref{fig:end_to_end_float32_latency_improvement} shows the latency speedup by using Tensor Cores(\texttt{NHWC}).
\texttt{TITAN\_V} achieves significant speedup (up to $1.72\times$).
We can see that \texttt{Tesla\_T4} benefits most from Tensor Cores for smaller batch sizes (i.e. might be best used for low-latency inference).

\subsection{\questionbox{\color{white} \small $\mathcal{Q}$\normalfont\sffamily\textsf{1,2,3,5,6}} Parallel Execution, Algorithm Selection, Layer Fusion, and Tensor Cores}\label{sec:tensorcore_fusion}

\dnnscope{} can be used to perform the above analysis jointly.
To demonstrate this, we analyzed the latency speedup when using parallel execution of data-independent layers, optimal algorithm selection, layer fusion, and Tensor Cores (\texttt{NHWC}).
Figure~\ref{fig:end_to_end_fused_nhwc_float32_latency_improvement} shows the latency speedup for \texttt{ResNet50-v1} across batch sizes and systems.
Up to a $1.95\times$ and $1.8\times$ speedup can be achieved by \texttt{TITAN\_V} and \texttt{Tesla\_V100} respectively.
We can surmise, from the previous analysis, that most of the profit for \texttt{TITAN\_V} is attributed to its use of Tensor Cores.
\texttt{Quadro\_RTX} and \texttt{Telsa\_T4} achieve marginal speedup over the Tensor Core results.

\section{Related Work}

\textit{DL Benchmarking}:
There has been no shortage of work on developing benchmarks to characterize DL models.
These DL benchmarks either take a model as a black-box and measure the user-observable latency and throughput (end-to-end benchmarks) or delve deeper into models to characterize the layer or kernel performance (micro-benchmarks).
The end-to-end benchmarks~\cite{mlperf,aimatrix,dawnbench} provide a corpus of models that are deemed to be of value to characterize for industry and research.
Micro-benchmarks~\cite{deepbench,convbench,benchdnn,aimatrix} distill DL models into their layers or kernels, and are hand-curated.
Micro-benchmarking enables easy measurements of layers within popular DL models and integrates easily with profiling tools.
In~\cite{xsp}, the author present a design that enables benchmarking DL models at across the abstraction levels of inference pipeline and introduce a hierarchical profiling methodology (enabling framework-, model-, and hardware-profiling).
In~\cite{deep500}, the authors propose a benchmark suite to enable fair comparison of DL techniques at different levels of granularity.
At the operator level, \cite{deep500} takes ONNX models and generates micro-benchmarks that target the framework's Python API to measure the latency of each operator.
\dnnscope{} also takes ONNX models as input, but generates lower-level cuDNN and cuBLAS micro-benchmarks to compute the ``lower-bound'' latency of the model, and perform analysis.
The authors are unaware of previous work which generates micro-benchmarks from model layers and couples it with an analysis workflow to inform optimizations.

\textit{Performance Advising}:
There is past work on using profiling to inform users of possible optimizations.
These optimizations are performed at the compiler level~\cite{ashouri2019survey} or are plugins to code editors to inform proper usage of APIs\cite{vandierendonck2010paralax,haj2019neurovectorizer}.
Low-level profile reports and some suggestions on how to address bottlenecks are provided by profilers  and IDEs such as: NVIDIA's Nvprof~\cite{nvprof}, Intel's VTune~\cite{vtune}, Oracle's Solaris Studio~\cite{solarisoracle}, Microsoft's Roslyn~\cite{ng2011roslyn}, and IBM's XL~\cite{du2015explore}.
To the author's knowledge, there has been no work on applying or specializing the optimization advising to the DL domain.

\section{Conclusion}\label{sec:conluation}

This paper presents \dnnscope, a sustainable and extensible DL benchmarking and analysis design that automatically generates layer-wise benchmarks for DL models to compute the ``lower-bound'' latency and inform optimizations on GPUs.
We use \dnnscope{} to evaluate a set of $30$ models using different frameworks on $7$ GPUs, and pinpointed the optimizations in parallel layer execution, cuDNN algorithm selection, framework inefficiency, layer fusion, and Tensor Core usage.
The results show that \dnnscope{} fills a significant gap within the characterization/optimization cycle and would boost the productivity of DL model, framework, and library developers.

\section*{Acknowledgments}
\label{sec:ack}

This work is supported by the IBM-ILLINOIS Center for Cognitive Computing Systems Research (C3SR) - a member of the IBM Cognitive Horizon Network, and the Applications Driving Architectures (ADA) Research Center - one of the JUMP Centers co-sponsored by SRC and DARPA.

\bibliography{IEEEabrv,main}

\end{document}